\documentclass{article} 
\ifdefined\pdfminorversion
\fi
\usepackage{iclr2027_conference,times}

\usepackage{amsmath,amsfonts,bm}

\def\eqref#1{equation~\ref{#1}}

\def\1{\bm{1}}

\DeclareMathAlphabet{\mathsfit}{\encodingdefault}{\sfdefault}{m}{sl}
\SetMathAlphabet{\mathsfit}{bold}{\encodingdefault}{\sfdefault}{bx}{n}

\usepackage{hyperref}
\hypersetup{hidelinks}
\usepackage{url}
\usepackage{booktabs}
\usepackage{algorithm}
\usepackage{algpseudocode}
\usepackage{graphicx}
\usepackage[caption=false,font=small]{subfig}
\usepackage[table]{xcolor}
\definecolor{tablegreen}{HTML}{2E8B57}
\definecolor{tableamber}{HTML}{9A86B5} 
\newcommand{\best}[1]{\cellcolor{tablegreen!20}\textbf{#1}}
\newcommand{\second}[1]{\cellcolor{tableamber!24}\underline{#1}}
\newcommand{\bestlabel}{\colorbox{tablegreen!20}{\textbf{Best}}}
\newcommand{\secondlabel}{\colorbox{tableamber!24}{\underline{second-best}}}

\usepackage{enumitem}
\usepackage{tcolorbox}
\newcommand{\promptfont}{\fontfamily{ptm}\selectfont\scriptsize}
\newcommand{\prompttitlefont}{\fontfamily{ptm}\selectfont\scriptsize\bfseries}

\usepackage{mathtools}
\usepackage{amsmath}
\usepackage{amssymb}

\title{Beyond Timestamps: Decision-Aligned On-Policy Distillation for Long-Horizon Agents}

\author{Mingju Chen$^{1,2*}$, Can Lv$^{1,2*}$, Jinrong Liu$^2$, Huan Zhang$^{1,2}$, \textbf{Heng Chang}$^{3}$, \textbf{Shiji Zhou}$^{1,2}$
\\[0.25em]
$^1$Beijing Advanced Innovation Center for Future Blockchain and Privacy Computing,\\ Beihang University\; $^2$ School of Artificial Intelligence, Beihang University\\
$^3$Tsinghua University 
$^*$ Equal Contribution \\
{\small
   \textbf{Project Lead:} Heng Chang,
   \textbf{Corresponding to:} Shiji Zhou \href{mailto:zhoushiji25@buaa.edu.cn}{\textless zhoushiji25@buaa.edu.cn\textgreater}
}\\[0.05em]}

\iclrfinalcopy
\begin{document}

\maketitle

\begin{abstract}

Reinforcement learning with verifiable rewards (RLVR) often relies on sparse outcome rewards, providing coarse supervision for long-horizon agents. On-policy self-distillation (OPSD) complements this signal with dense privileged feedback. However, we identify \emph{Decision--Timestamp Mismatch}: privileged guidance may be misaligned with the student's functional decision because the corresponding decision can occur at a different timestep, while the student's decision itself may span multiple timesteps rather than being tied to a single timestamp. Thus, timestamp-local supervision can misalign both the context and the temporal scope of credit. To address this mismatch, we introduce \textsc{AlignOPSD}, following the principle of aligning supervision before assigning credit. Decision-Aligned Supervision Rectification re-scores the same student-sampled response in functionally matched contexts across sibling rollouts to calibrate local teacher evidence. Semi-Markov Hierarchical Credit Assignment then derives variable-duration decision spans from correspondence changes and uses rectified evidence to allocate outcome-grounded credit across spans and their constituent turns. We evaluate \textsc{AlignOPSD} with Qwen2.5-3B and Qwen2.5-7B on ALFWorld, WebShop, and Search-QA against representative baselines. \textsc{AlignOPSD} outperforms both GRPO and StepOPSD across all eight backbone--aggregate-metric comparisons, improving on GRPO by 5.5--8.7 \% and ranking first in six. Additional analyzes examine the two alignment stages and hyperparameter sensitivity between tasks. Our code is avaliable at \url{https://github.com/mingju-c/Align-OPSD}.

\end{abstract}

\section{Introduction}
\label{Intro}

Agentic long-horizon tasks require language agents to accomplish goals through a sequence of interdependent intermediate decisions, while training rewards are often available only at the end of a task. Methods such as Group Relative Policy Optimization (GRPO) compare terminal rewards across sibling rollouts and broadcast a trajectory-level advantage to the generated tokens \citep{shao2024deepseekmathpushinglimitsmathematical}. Such supervision captures overall task performance, but cannot directly distinguish effective decisions from incidental behaviors within successful trajectories, nor can it precisely localize errors within failed ones \citep{NEURIPS2025_420c9f77,zhang2026trcatransitionwiserubriccredit}. To complement this coarse-grained supervision, on-policy distillation (OPD) and its self-distilled variants (OPSD) provide dense teacher feedback on behaviors generated by the student policy itself \citep{ICLR2024_5be69a58,zhao2026selfdistilledreasoneronpolicyselfdistillation}. When the teacher is further granted privileged task information, such feedback can provide richer local evidence about intermediate behaviors \citep{penaloza2026privilegedinformationdistillationlanguage,lu2026selfdistilledagenticreinforcementlearning}. The challenge is therefore not merely to obtain denser supervision, but to determine which teacher evidence can meaningfully inform the student's intermediate decisions at different stages of execution.

Toward more effective intermediate supervision, recent agentic OPSD methods have developed along two main directions. One line enriches teacher evidence through hindsight information, execution feedback, or privileged contexts \citep{lu2026selfdistilledagenticreinforcementlearning,wang2026tcodexploringtemporalcurriculum,yu2026multirolloutonpolicydistillationpeer,yang2026opidonpolicyskilldistillation,zhang2026latentonpolicyselfdistillation}, while another refines how teacher--student discrepancies are filtered and aggregated into action- or turn-level learning signals \citep{zhang2026stepopsdstepawareonlinepreference,li2026geargranularityadaptiveadvantagereweighting}. These advances improve the informativeness and utilization of supervision, yet their local signals remain anchored to corresponding positions along the student trajectory. Interpreting these local discrepancies as decision-level credit relies on a key assumption: \textit{teacher evidence at the current trajectory position provides an appropriate basis for evaluating the functional decision pursued by the student at that position.} However, these refinements do not guarantee functional correspondence between teacher and student behaviors across rollouts, timestamps, and decision boundaries.

\begin{figure}[t]
    \centering
    \includegraphics[width=\linewidth]{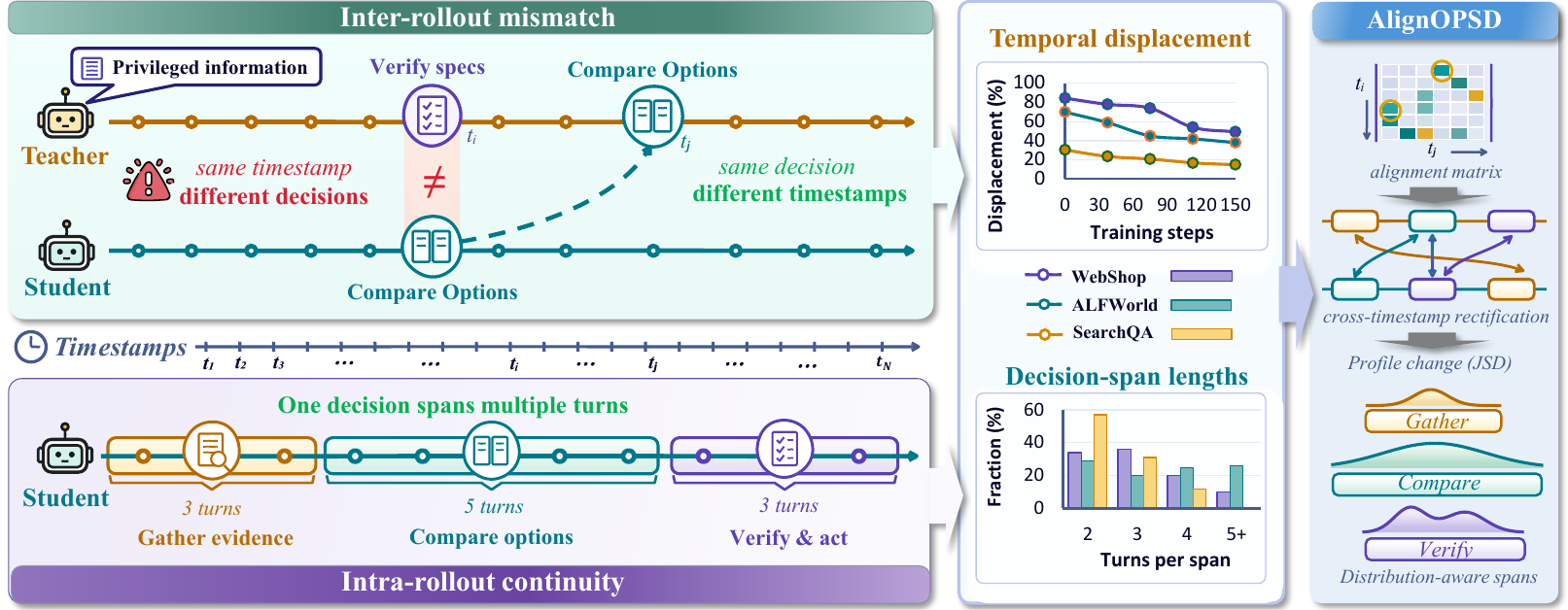}
    \vspace{-15 pt}
    \caption{\textbf{Decision--Timestamp Mismatch.} Functional decisions need not share timestamps across different sibling rollouts and can span multiple turns within individual trajectories.}
    \label{fig:intro1}
    \vspace{-18pt}
\end{figure}

We refer to this phenomenon as \textit{Decision--Timestamp Mismatch}, which manifests at two related levels. Across rollouts, the same timestamp need not identify
the same decision context, while corresponding decisions may occur at different
turns. As illustrated in Figure~\ref{fig:intro1}, the
student compares product options while the privileged teacher verifies
specifications, the corresponding comparison appears later in the teacher
rollout. Even under the same visible history, privileged information may change
which decision the teacher favors. Consequently, although same-prefix
probability comparisons remain valid, the resulting disagreement need not
isolate the quality of the student's current decision. The issue is therefore the functional comparability of supervisory contexts for reliable decision-level credit attribution, not the validity of same-prefix probability comparisons.

Within a rollout, a functional decision may persist across multiple
turns, with boundaries that do not coincide with individual timestamps. Assigning credit independently to each turn can fragment
behaviors that jointly realize one decision, whereas fixed-length windows can
mix distinct decisions. The upper-right panels illustrate how these two aspects
can be characterized quantitatively through the temporal displacement of
corresponding decisions and the distribution of decision-span lengths.
Motivated by this, we study: \textbf{\textit{How can functional correspondence guide both the selection of privileged evidence and the temporal scope of outcome-grounded credit assignment?}}

To address this problem, we propose \textsc{AlignOPSD}, which aligns privileged supervision with functional decisions before assigning credit. Specifically, Decision-Aligned Supervision Rectification estimates functional correspondences across sibling rollouts, re-scores the same student-sampled response in matched privileged contexts, and combines cross-context and local evidence according to correspondence confidence. Building on this alignment, Semi-Markov Hierarchical Credit Assignment constructs variable-duration decision spans from correspondence changes, allocates outcome-grounded credit across spans and their constituent turns using rectified evidence, and broadcasts turn-level credit to the corresponding training tokens. We evaluate \textsc{AlignOPSD} on ALFWorld, WebShop, and Search-QA under matched training and evaluation protocols against representative agentic RL methods spanning embodied, search, and shopping environments.

Our contributions are summarized as follows:

\begin{itemize}[leftmargin=*, itemsep=5pt, parsep=0pt]

\item \textbf{Problem Identification.}
We identify \textit{Decision--Timestamp Mismatch}: timestamp-local
privileged evidence may not correspond to the student's functional
decision, while the same decision can occur at positions across
rollouts, leading to mismatched supervisory contexts and credit
horizons.

\item \textbf{Proposed Solution.}
We propose \textsc{AlignOPSD}, which establishes functional
correspondence before credit assignment by rectifying privileged
supervision with matched contexts and deriving adaptive decision spans
for outcome-grounded credit allocation over variable-duration decisions.

\item \textbf{Experimental Validation.}
Across three agentic benchmarks, \textsc{AlignOPSD} improves over GRPO
by 5.5--8.7 \% in all eight backbone--metric comparisons and ranks first
in six. Additional analyses characterize correspondence alignment,
span adaptation, and supervision rectification.

\end{itemize}

\section{Methodology}
\label{sec:method}

\subsection{Problem Formulation}
\label{sec:formulation}

\begin{figure}[t]
    \centering
    \includegraphics[width=\linewidth]{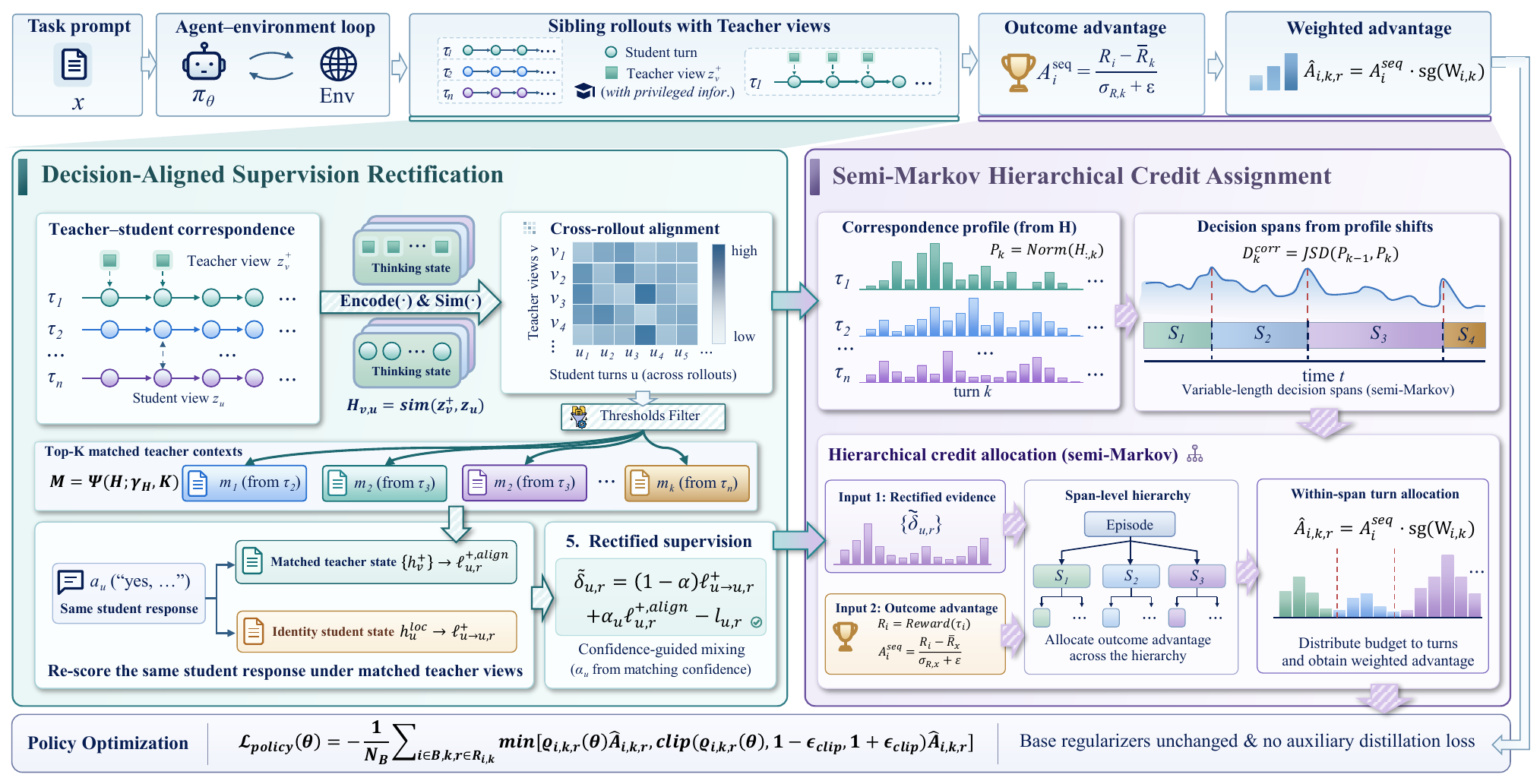}
    \vspace{-13pt}
    \caption{\textbf{Overview of \textsc{AlignOPSD}.} Cross-rollout alignment establishes comparable supervisory contexts, while correspondence changes define decision spans for hierarchical credit assignment. The weights enter policy optimization, with auxiliary components used during training.}
    \label{fig:intro}
    \label{fig:overview}
    \vspace{-5pt}
\end{figure}

Given task \(x\) and initial observation \(o_0\), trajectory \(i\) starts from
\(h_{i,1}=(x,o_0)\). At turn \(k\),
\begin{equation}
a_{i,k}=(y_{i,k,r})_{r=1}^{L_{i,k}}
\sim\pi_\theta(\cdot\mid h_{i,k}),
\qquad
h_{i,k+1}=h_{i,k}\oplus(a_{i,k},o_{i,k}),
\end{equation}
where \(a_{i,k}\) may contain a thinking trace followed by an executable action,
\(o_{i,k}\) is the observation returned by the environment, and \(\oplus\)
appends the response--observation pair to the interaction history. After \(K_i\)
turns, trajectory \(\tau_i\) receives a verifiable terminal reward
\(R_i=R(\tau_i)\).
For each task \(x\), group-relative policy optimization samples \(G\) sibling
trajectories \(\mathcal G_x\) and computes
\begin{equation}
A_i^{\mathrm{seq}}
=
\frac{R_i-\overline R_x}{\widehat\sigma_{R,x}+\epsilon},
\qquad
\overline R_x=\frac{1}{G}\sum_{j\in\mathcal G_x}R_j .
\end{equation}
The base objective is broadcast \(A_i^{\mathrm{seq}}\) to the optimized response
tokens of the trajectory \(i\), providing a scale and direction update based on
the results. In OPSD-style training, each interaction history \(h_{i,k}\) is paired with a
privileged view \(h^+_{i,k}=(h_{i,k},k_x)\), where \(k_x\) is task-relevant privileged information obtained through relevance-based retrieval and available only during training. Both views share the same
policy \(\pi_\theta\). We denote the ordinary student view by
\(\pi_\theta(\cdot\mid h_{i,k})\) and define the privileged teacher view as
\(
\pi_\theta^+(\cdot\mid h_{i,k})
\triangleq
\pi_\theta(\cdot\mid h^+_{i,k}).
\)
Thus, the teacher differs from the student only in its conditioning context,
rather than its model parameters, as shown in Figure~\ref{fig:overview}.

\subsection{Decision-Aligned Supervision Rectification}
\label{sec:rectification}

For compact notation, let \(u=(i,k)\) denote a target turn and
\(v=(j,l)\) a candidate source turn from another sibling rollout.
Conventional OPSD evaluates \(a_u\) only under its timestamp-local privileged
view \(h_u^+\), which may be misaligned with the functional decision represented
at \(u\). We therefore establish cross-rollout decision correspondence before
rectifying the local teacher evidence.

\textbf{Decision-state correspondence.}
Decision-relevant information may be distributed across observations and
interaction history, making direct state matching difficult. We therefore use
thinking traces as semantic surrogates of the decision state. Let \(z_u\) denote
the thinking parsed from student response \(a_u\), and \(z_v^+\) the privileged
thinking generated by the teacher view \(\pi_\theta^+(\cdot\mid h_v)\).
A frozen semantic encoder \(\mathbf{Enc(\cdot)}\) maps both into a shared decision-state
space:
\begin{equation}
\mathbf d_u=\mathbf{Enc}(z_u),
\qquad
\mathbf d_v^+=\mathbf{Enc}(z_v^+).
\end{equation}
We treat proximity in this space as approximate functional-decision
correspondence. Appendix~\ref{app:thinking} audits this surrogate with a
method-faithful comparison of exact same-state and cross-rollout pairs.

\textbf{Decision-alignment operator.}
Given the student and teacher-side decision-state representations, we construct
a source-by-target correspondence matrix 
\begin{equation}
H^{(n)}_{v,u}
=
\cos\!\left(
\mathbf d_v^{+,(n)},
\mathbf d_u^{(n)}
\right),
\qquad
\mathcal M^{(n)}
=
\Psi\!\left(H^{(n)};\gamma_H,K\right).
\end{equation}
For each target turn \(u\), \(\Psi\) considers structurally valid source turns
from other rollouts of the same task, removes candidates below similarity
threshold \(\gamma_H\), retains at most one source per sibling rollout, and
selects the \(top\text{-}K\) matches. Across all three benchmarks, an
action-consistency filter requires valid parsed actions and an action-embedding
cosine of at least 0.8. On WebShop, where actions admit exact symbolic checks,
we additionally require matching operation types and identical click targets.
Each nonzero column
\(\mathcal M^{(n)}_{\cdot,u}\) assigns normalized weights to the retained
reference views, while unmatched targets receive a zero column. The operator is
recomputed for each on-policy batch, allowing the correspondence structure to
evolve with the policy during training. 

\textbf{Multi-view supervision rectification.}
Instead of corresponding to the identity view \(v=u\), our alignment
operator additionally provides off-diagonal reference views for the same target
response. Rather than imitating retrieved responses, the privileged policy $\pi_\theta^+$
teacher-forces the same complete response \(a_u\) under each aligned
source history from a sibling rollout:
\begin{equation}
\ell^+_{v\rightarrow u,r}
=
\log\pi_\theta^+(y_{u,r}\mid h_v,y_{u,<r}),
\qquad
\ell_{u,r}
=
\log\pi_\theta(y_{u,r}\mid h_u,y_{u,<r}).
\end{equation}
The identity view gives the conventional local gap
$
\delta^{\mathrm{id}}_{u,r}
=
\ell^+_{u\rightarrow u,r}-\ell_{u,r}.
$
For a matched target, \(\mathcal M^{(n)}_{\cdot,u}\) aggregates the aligned
privileged views and rectifies the local evidence as
\begin{equation}
\ell^{+,\mathrm{align}}_{u,r}
=
\log\!\sum_v
\mathcal M^{(n)}_{v,u}
\exp~\!\bigl(\ell^+_{v\rightarrow u,r}\bigr),\quad
\widetilde\delta_{u,r}
=
(1-\alpha_u)\ell^+_{u\rightarrow u,r}
+
\alpha_u\ell^{+,\mathrm{align}}_{u,r}
-
\ell_{u,r}.
\end{equation}
The coefficient \(\alpha_u\in[0,\alpha_{\max}]\) adaptively controls how
strongly aligned views modify the local signal according to the quality of the match. Concretely, we
compute
\begin{equation}
\widehat H_{v,u}
=
\operatorname{clip}\!\left(
\frac{H_{v,u}-\gamma_H}{1-\gamma_H},
0,1
\right),
\qquad
\rho_u
=
\sum_v \mathcal M^{(n)}_{v,u}\widehat H_{v,u},\qquad\alpha_u=\alpha_{\max}\rho_u
\end{equation}
If no valid source is retrieved, we set \(\alpha_u=0\), reducing
\(\widetilde\delta_{u,r}\) to the conventional identity gap
\(\delta^{\mathrm{id}}_{u,r}\). The rectified gap
\(\widetilde\delta_{u,r}\) is then passed to the hierarchical credit allocator
in Section~\ref{sec:credit_assignment}.

\subsection{Semi-Markov Hierarchical Credit Assignment}
\label{sec:credit_assignment}

Rectification provides decision-relevant supervisory references, but a functional decision may span multiple interaction turns. Thus, we organize credit over variable-duration decision spans and introduce a hierarchical allocator \(\Phi\) that converts rectified evidence into outcome-grounded advantages.

\textbf{Correspondence-driven spans.}
A continuing functional decision is expected to retain similar reference
contexts across adjacent turns. We normalize correspondence scores over
structurally valid sources from other same-task rollouts before top-\(K\)
truncation:
\begin{equation}
\mathbf P^{(n)}_{i,k}
=
\operatorname{Norm}\!\left(H^{(n)}_{\cdot,(i,k)}\right),
\qquad
D^{\mathrm{corr}}_{i,k}
=
\operatorname{JSD}\!\left(
\mathbf P^{(n)}_{i,k-1},
\mathbf P^{(n)}_{i,k}
\right).
\end{equation}
where \(D^{\mathrm{corr}}_{i,k}\) measures the correspondence shift between
adjacent turns. A batch-adaptive threshold and duration constraints produce a contiguous partition
\(\mathcal S_i=\{S_{i,1},\ldots,S_{i,J_i}\}\), whose variable-duration spans
serve as credit units under a semi-Markov abstraction.

\textbf{Outcome-oriented evidence.}
The rectified gap provides token-level teacher support, while hierarchical
allocation operates at the turn level. We therefore define an outcome-oriented
evidence score measuring agreement between rectified support and the
trajectory-level outcome direction:
\begin{equation}
\mathcal E_{i,k}
=
\operatorname{sgn}\!\left(A_i^{\mathrm{seq}}\right)
\frac{1}{N_{i,k}}
\sum_{r=1}^{N_{i,k}}
\widetilde\delta_{i,k,r}
\end{equation}
where \(N_{i,k}\) denote the number of optimized response tokens at turn \(k\), a larger \(\mathcal E_{i,k}\) indicates stronger outcome-consistent teacher
support and therefore stronger evidence for allocating credit to turn \(k\).
For span \(S_{i,m}\), we define
\(\mathcal E^{\mathrm{span}}_{i,m}\) as the mean evidence over all turns in the corresponding span.

\textbf{Hierarchical credit allocation.}
We view \(A_i^{\mathrm{seq}}\) as an outcome-grounded credit budget and use the
decision hierarchy to redistribute it. Let
\(\mathbf E_i=(\mathcal E_{i,k})_k\) denote the turn-level evidence, and define
the cumulative credit measure
\(\Lambda_{i,k}=\sum_{t\leq k}N_{i,t}\), whose increments recover the
uniform per-token allocation of the objective. We introduce a credit allocator $\Phi$ to produce turn-level credit weights:
\begin{equation}
\mathbf W_i
=
\Phi\!\left(
\mathcal S_i,\mathbf E_i;\Lambda_i
\right).
\end{equation}
Internally, \(\Phi\) performs two-level allocation. For each span \(S_{i,m}\), its aggregate evidence \(\mathcal E^{\mathrm{span}}_{i,m}\) determines a span-level budget \(B_{i,m}\). Conditioned on this budget, turn-level evidence determines the within-span allocation \(Q_{i,k\mid m}\):
\begin{equation}
B_{i,m}
=
\Phi_{\mathrm{span}}
\!\left(
\mathcal E^{\mathrm{span}}_{i,m},
\Delta\Lambda_{i,m}
\right),
\qquad
W_{i,k}
=
\operatorname{Norm}\!\left[
B_{i,m}\,
\Phi_{\mathrm{turn}}
\!\left(
\mathcal E_{i,k},
\Delta\Lambda_{i,k}
\mid S_{i,m}
\right)
\right].
\end{equation}
Thus, the outer allocation assigns span-level credit, while the inner allocation
distributes it across constituent turns. After bounded normalization, turn
weights reweight the trajectory advantage:
\begin{equation}
\widehat A_{i,k}
=
A_i^{\mathrm{seq}}
\operatorname{sg}(W_{i,k}),
\end{equation}
where \(\operatorname{sg}\) denotes stop-gradient. KL regularization constrains
both allocation levels from deviating excessively from the neutral credit
measure, while normalization preserves the token-weighted mean advantage.
Exact allocation and normalization details are provided in
Appendix~\ref{app:invariants}.

\textbf{Policy optimization.}
Each optimized token inherits its turn advantage, \(\widehat A_{i,k,r}=\widehat A_{i,k}\). For the current trajectory batch \(\mathcal B\), let \(\pi_{\theta_{\mathrm{old}}}\) denote the fixed snapshot used for rollout collection and evidence construction. Define the policy ratio and total optimized-token count as
\begin{equation}
\varrho_{i,k,r}(\theta)
=
\frac{\pi_\theta(y_{i,k,r}\mid h_{i,k},y_{i,k,<r})}
{\pi_{\theta_{\mathrm{old}}}(y_{i,k,r}\mid h_{i,k},y_{i,k,<r})},
\qquad
N_{\mathcal B}=\sum_{i\in\mathcal B}\sum_k N_{i,k}.
\end{equation}
The token-mean clipped policy loss is
\begin{equation}
\begin{aligned}
\mathcal L_{\mathrm{policy}}(\theta)
={}&
-\frac{1}{N_{\mathcal B}}
\sum_{\substack{i\in\mathcal B,\,k\\r\in\mathcal R_{i,k}}}
\min\!\Bigl[
\varrho_{i,k,r}(\theta)\widehat A_{i,k,r},
\operatorname{clip}\!\left(\varrho_{i,k,r}(\theta),1-\epsilon_{\mathrm{clip}},1+\epsilon_{\mathrm{clip}}\right)\widehat A_{i,k,r}
\Bigr],
\end{aligned}
\end{equation}
where \(\epsilon_{\mathrm{clip}}\) is the base clipping threshold. Other base-objective regularization terms remain unchanged, and no auxiliary distillation loss is added. All correspondence, span, evidence, and allocation quantities are stop-gradient and used only during training.

\section{Experiments}
\label{sec:experiments}


\subsection{Experimental Setup}
\label{sec:protocol}

\begin{table}[!t]
    \centering
    \caption{
        \textbf{Performance on ALFWorld, Search-QA, and WebShop.}
        We report success rate (\%) on ALFWorld, accuracy (\%) on Search-QA,
        and Score/Acc (\%) on WebShop. Skills are training-only unless marked
        with $*$ (validation with skills).
        \bestlabel{} and \secondlabel{} are highlighted.
    }
    \label{tab:main_results}

    \resizebox{\textwidth}{!}{%
    \begin{tabular}{l ccccccc cccccccc cc}
    \toprule
    & \multicolumn{7}{c}{\textbf{ALFWorld}}
    & \multicolumn{8}{c}{\textbf{Search-QA}}
    & \multicolumn{2}{c}{\textbf{WebShop}} \\
    \cmidrule(lr){2-8}
    \cmidrule(lr){9-16}
    \cmidrule(lr){17-18}
    \textbf{Method}
    & \textbf{Pick} & \textbf{Heat} & \textbf{Look} & \textbf{Clean}
    & \textbf{Cool} & \textbf{Pick2} & \textbf{Avg}
    & \textbf{NQ} & \textbf{Triv} & \textbf{Pop} & \textbf{Hotp}
    & \textbf{2Wk} & \textbf{MuS} & \textbf{Bam} & \textbf{Avg}
    & \textbf{Score} & \textbf{Acc} \\
    \midrule

    \rowcolor{gray!10}
    \multicolumn{18}{l}{\textit{Qwen2.5-3B-Instruct}} \\
Vanilla
        & 44.4 & 15.4 & 11.1 & 6.2 & 28.6 & 12.5 & 21.9 & 24.6 & 48.1 & 31.0 & 26.3 & 25.3 & 7.2 & 59.7 & 31.7 & 6.7 & 0.8 \\
Skill-Prompt$^*$
        & 51.7 & 0.0 & 66.7 & 48.4 & 4.3 & 10.0 & 28.9 & 23.7 & 46.2 & 30.6 & 24.4 & 22.1 & 7.5 & 12.5 & 23.9 & 1.2 & 0.8 \\
OPSD
        & 48.8 & 0.0 & 41.7 & 16.7 & 15.8 & 16.7 & 28.1 & 0.1 & 0.1 & 0.1 & 0.0 & 0.0 & 0.0 & 0.0 & 0.0 & 11.3 & 3.1 \\
GRPO
        & 91.2 & 61.9 & 62.5 & \second{96.2} & 65.0 & 47.4 & 75.0 & 39.3 & 60.6 & 41.1 & 37.4 & 34.6 & 15.4 & 26.4 & 36.4 & 79.8 & 63.3 \\
Skill-GRPO
        & 88.9 & 70.6 & 71.4 & 58.8 & 40.7 & 29.2 & 60.2 & 43.5 & 58.8 & 43.0 & 36.8 & 32.2 & 11.7 & 12.5 & 34.1 & 77.3 & 60.9 \\
Skill-GRPO$^*$
        & 94.3 & 66.7 & 57.1 & \best{100} & 73.1 & 57.1 & \second{80.5} & 44.3 & 59.6 & 44.3 & 39.0 & 36.1 & 14.5 & 14.9 & 36.1 & 76.3 & \second{66.4} \\
GRPO+OPSD
        & \best{100} & \second{75.0} & \best{82.4} & 85.7 & 70.0 & 60.0 & \best{81.2} & 44.3 & \second{60.8} & \second{46.0} & 38.9 & 38.6 & 13.9 & 64.5 & 43.9 & 77.8 & \second{66.4} \\
Skill-SD
        & 88.2 & 52.4 & 50.0 & \second{96.2} & 65.0 & 57.9 & 73.4 & 42.8 & 60.1 & 42.2 & \best{39.5} & \second{38.9} & 13.6 & 64.9 & 44.0 & 75.9 & 64.0 \\
RLSD
        & 87.9 & \second{75.0} & \second{75.0} & 90.9 & 73.1 & 68.4 & 79.7 & 42.3 & 58.6 & 43.3 & \best{39.5} & \best{39.6} & \best{16.1} & \best{66.1} & \second{44.2} & \second{84.4} & \second{66.4} \\
SDAR
        & \second{94.6} & 50.0 & 66.7 & 68.2 & 60.0 & \best{80.0} & 74.2 & \best{44.8} & 58.1 & 44.3 & 38.6 & 36.2 & 15.7 & \best{66.1} & 43.4 & 70.7 & 57.8 \\
StepOPSD
        & 82.4 & 52.2 & 66.7 & 82.6 & \second{73.7} & \second{75.0} & 73.4 & 43.6 & \best{61.2} & 43.8 & \second{39.2} & 38.1 & \second{15.8} & 64.5 & 43.7 & 82.4 & \second{66.4} \\
\textbf{\textsc{AlignOPSD}}
        & 88.6 &  \best{81.8} & 54.5 & 90.3 & \best{82.6} & 58.8 & \second{80.5} & \second{44.5} & \best{61.2} & \best{46.3} & \best{39.5} & \second{38.9} & 13.7 & \second{65.3} & \best{45.1} & \best{86.8} & \best{68.8} \\

    \midrule
    \rowcolor{gray!10}
    \multicolumn{18}{l}{\textit{Qwen2.5-7B-Instruct}} \\

Vanilla
        & 36.1 & 0.0 & 22.2 & 3.1 & 0.0 & 0.0 & 12.5 & 25.2 & 50.8 & 29.5 & 29.0 & 29.0 & 10.4 & 63.7 & 33.9 & 5.9 & 1.6 \\
Skill-Prompt$^*$
        & 51.7 & 5.3 & 50.0 & 32.3 & 4.3 & 0.0 & 23.4 & 30.9 & 52.1 & 32.7 & 32.7 & 27.9 & 12.7 & 66.1 & 36.4 & 1.7 & 0.8 \\
OPSD
        & 50.0 & 21.4 & 60.0 & 22.7 & 17.6 & 9.5 & 32.8 & 8.8 & 8.6 & 17.5 & 2.5 & 4.2 & 0.5 & 1.2 & 6.2 & 4.5 & 2.3 \\
GRPO
        & 91.2 & 81.0 & \second{87.5} & 96.2 & 65.0 & 57.9 & 81.2 & 45.1 & 63.7 & 44.0 & 43.6 & 43.2 & 16.8 & 37.6 & 42.0 & 80.9 & 72.6 \\
Skill-GRPO
        & 88.5 & 61.1 & 66.7 & 65.2 & 57.7 & \second{73.1} & 69.5 & 45.2 & 63.7 & 45.7 & 43.1 & 43.3 & 19.6 & 21.4 & 40.3 & 80.4 & 71.9 \\
Skill-GRPO$^*$
        & \best{100} & 83.3 & 83.3 & \second{96.4} & 75.0 & \textbf{78.9} & \second{88.4} & 44.8 & 63.0 & 45.1 & 43.7 & 43.7 & \second{20.5} & \second{71.4} & 47.5 & 87.0 & \best{81.2} \\
GRPO+OPSD
        & 91.4 & 87.5 & 61.5 & \best{100} & \second{76.5} & 52.2 & 80.4 & \best{47.3} & 64.5 & 46.9 & 43.8 & 39.3 & 18.0 & 69.4 & 47.0 & 86.8 & 76.5 \\
Skill-SD
        & 93.9 & \best{100} & \best{93.8} & 90.9 & 69.2 & 68.4 & 85.1 & \second{47.1} & 64.5 & 47.8 & 44.2 & 42.1 & 20.2 & 69.0 & 47.8 & 86.1 & 76.5 \\
RLSD
        & \best{100} & 58.8 & \second{87.5} & 92.3 & \best{80.0} & 65.2 & 82.0 & 46.8 & 63.0 & 44.4 & \best{45.5} & \best{48.9} & \best{21.5} & \best{73.0} & \second{49.0} & \second{87.4} & 77.3 \\
SDAR
        & 94.7 & 86.7 & 75.0 & \best{100} & 68.2 & \textbf{78.9} & 85.9 & 46.3 & 63.5 & \second{48.2} & 43.8 & \second{48.4} & 19.6 & \best{73.0} & \second{49.0} & 83.6 & 74.2 \\
StepOPSD
        & \second{98.1} & \second{90.5} & 75.0 & \best{100} & \best{80.0} & 63.2 & \second{88.4} & 45.3 & \second{64.6} & 45.1 & 44.5 & 44.4 & 19.3 & 69.8 & 48.2 & 87.2 & 78.1 \\
\textbf{\textsc{AlignOPSD}}
        & 70.8 & \best{100} & 81.8 & \best{100} & \second{76.5} & 70.8 & \best{89.1} & 46.4 & \best{65.6} & \best{48.5} & \second{44.6} & 43.7 & 19.5 & 69.8 & \best{49.1} & \best{87.9} & \second{78.9} \\
    \bottomrule

    \end{tabular}%
    }
    \vspace{-12 pt}
\end{table}

\textbf{Benchmarks.}
We evaluate on ALFWorld \citep{shridhar2021alfworldaligningtextembodied}, Search-QA
\citep{jin2025searchr1trainingllmsreason}, and WebShop \citep{NEURIPS2022_82ad13ec}, covering embodied
household control, search-augmented question answering, and interactive online
shopping, respectively. We report success rate for ALFWorld, exact-match
accuracy for Search-QA, and normalized score and exact success for WebShop.
Dataset composition, splits, and per-category evaluation details are provided
in Appendix~\ref{app:benchmarks}.

\textbf{Baselines.}
We compare three groups: training-free prompting methods (Vanilla and
Skill-Prompt$^*$), RLVR methods (GRPO, Skill-GRPO, and Skill-GRPO$^*$), and
self-distillation RL methods (OPSD, GRPO+OPSD, Skill-SD, RLSD, SDAR, and
StepOPSD) \citep{shao2024deepseekmathpushinglimitsmathematical,zhao2026selfdistilledreasoneronpolicyselfdistillation,wang2026skillsdskillconditionedselfdistillationmultiturn,yang2026selfdistilledrlvr,lu2026selfdistilledagenticreinforcementlearning,zhang2026stepopsdstepawareonlinepreference}.
All skill-conditioned methods use the same SkillBank and retrieval source. Detailed configurations are given in Appendix~\ref{app:baselines}.

\textbf{Implementation details.}
We use Qwen2.5-3B-Instruct and Qwen2.5-7B-Instruct backbones
\citep{qwen2025qwen25technicalreport}. Training uses a single node with up to eight NVIDIA
A800 GPUs. Rollout settings, hyperparameters, and training diagnostics are
reported in Table~\ref{tab:protocol} and Appendices~\ref{app:protocol}--\ref{app:training_diagnostics}.

\subsection{Main Results}
\label{sec:main_results}

\textbf{Aggregate performance.} Table~\ref{tab:main_results} compares \textsc{AlignOPSD} with trajectory-level RLVR and OPSD+RLVR baselines under the matched evaluation protocol. \textsc{AlignOPSD} achieves 80.5\%/89.1\% average success on ALFWorld, 45.1\%/49.1\% accuracy on Search-QA, and 86.8\%/87.9\% Score on WebShop for the 3B/7B backbones, respectively, under both model scales. Across the eight backbone--aggregate-metric comparisons, \textsc{AlignOPSD} ranks first in six and second in two, surpassed only by GRPO+OPSD on 3B ALFWorld (81.2\%) and Skill-GRPO$^*$ on 7B WebShop Acc (81.2\%).

\textbf{Improvement over trajectory- and step-level credit.} Compared with GRPO, \textsc{AlignOPSD} improves all eight aggregate comparisons, by 5.5\%/7.9\% on ALFWorld, 8.7\%/7.1\% on Search-QA, 7.0\%/7.0\% on WebShop Score, and 5.5\%/6.3\% on WebShop Acc for the 3B/7B backbones. \textsc{AlignOPSD} also outperforms StepOPSD across all eight comparisons, showing that the gains extend beyond step-level supervision alone. Since skill-conditioned baselines share the same source, the results suggest that privileged information alone is insufficient. Behavioral alignment and outcome-conditioned credit assignment also matter. We next ablate the two alignment stages.

\begin{table*}[t]
\centering
\setlength{\belowcaptionskip}{4pt}
\caption{\textbf{Ablation Study.} Final Score and Acc for \textsc{AlignOPSD} and four ablation variants are reported on the left and success-rate curves over 150 training steps are shown on the right.}
\label{tab:ablation_variants}

\begin{minipage}{\textwidth}
\begin{minipage}[t]{0.49\linewidth}
\vspace{3pt}
\raggedright
\small
\setlength{\tabcolsep}{2.5pt}
\renewcommand{\arraystretch}{1.45}
\resizebox{\linewidth}{!}{%
\begin{tabular}{@{}clrr@{}}
\toprule
\# & Variant & Score & Acc \\
\midrule
-- & \textbf{\textsc{AlignOPSD}} & \textbf{87.9} & \textbf{78.9} \\
\midrule
1 & w/o rectification + adaptive allocation & 84.2 & 71.9 \\
2 & w/ rectification + token allocation & 81.6 & 71.1 \\
3 & w/ rectification + turn allocation & 86.3 & 77.3 \\
4 & w/ rectification + random allocation & 78.9 & 69.5 \\
\bottomrule
\end{tabular}%
}
\end{minipage}\hfill
\begin{minipage}[t]{0.49\linewidth}
\vspace{0pt}
\centering
\includegraphics[width=\linewidth]{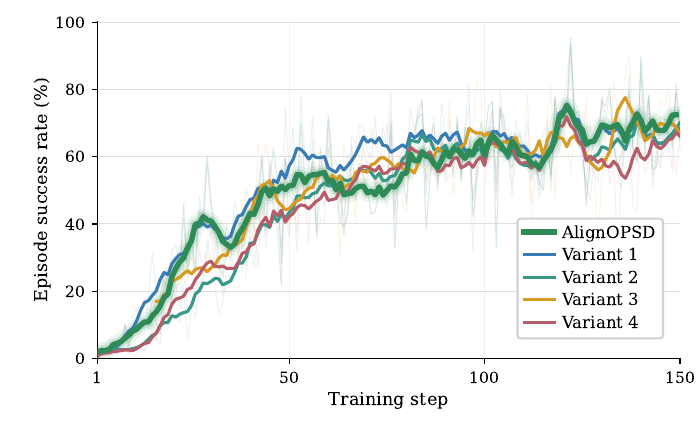}
\end{minipage}
\end{minipage}
\vspace{-25pt}
\end{table*}

\subsection{Ablation Study}
\label{sec:ablation}

\textbf{Effect of supervision rectification.}
Table~\ref{tab:ablation_variants} reports WebShop results with the 7B backbone. Removing Decision-Aligned Supervision Rectification while retaining adaptive allocation reduces Score from 87.9\% to 84.2\% and Acc from 78.9\% to 71.9\%, corresponding to drops of 3.7 and 7.0\%. This confirms that establishing a comparable supervisory context contributes beyond the allocation mechanism. The larger degradation in Acc indicates that rectification is particularly important for converting local teacher--student discrepancies into credit that supports complete task success.

\textbf{Effect of adaptive allocation.}
Keeping rectification fixed, we replace adaptive decision spans with token-, turn-, or random span-level allocation. Across both metrics, turn-level allocation outperforms token-level allocation, which in turn outperforms random allocation. Relative to these alternatives, \textsc{AlignOPSD} improves Score/Acc by 6.3/7.8\% over token-level allocation, 1.6/1.6\% over turn-level allocation, and 9.0/9.4\% over random allocation. These results show that turns already provide meaningful units for credit assignment, whereas uniformly finer token-level credit does not yield better supervision. The additional gain over turn-level allocation suggests that decision boundaries should adapt to correspondence changes during policy optimization rather than remain fixed.

\begin{figure*}[t]
\centering
\includegraphics[width=0.99\textwidth]{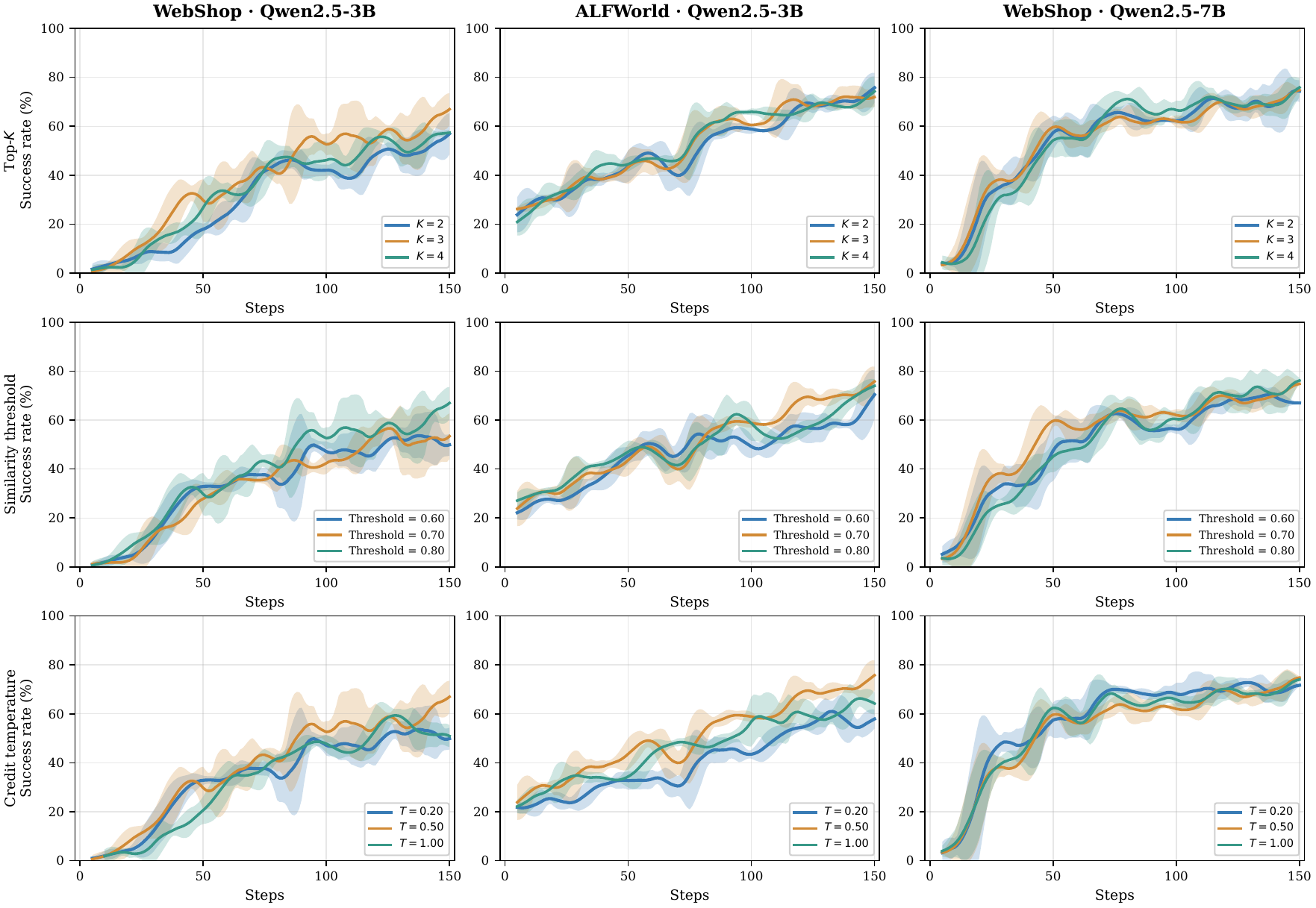}
\caption{\textbf{Sensitivity Analysis.} Sensitivity to retrieval top-$K$, thinking-similarity threshold $\gamma_H$, and credit temperature $T$ across the tested hyperparameter ranges. Curves smooth recorded validation checkpoints, and shading shows twice the local temporal variation rather than across-seed uncertainty.}
\label{fig:sensitivity_grid}
\vspace{-14pt}
\end{figure*}

\subsection{Sensitivity Analysis}
\label{sec:sensitivity}

Figure~\ref{fig:sensitivity_grid} examines sensitivity to retrieval top-$K$, thinking-similarity threshold $\gamma_H$, and credit temperature $T$, which control evidence coverage, correspondence filtering, and credit concentration, respectively. All panels report validation success rate from development histories, the curves characterize empirical operating regions rather than across-seed uncertainty.

\textbf{Retrieval top-$K$.}
The retrieval top-$K$ controls retrieval breadth, balancing broader evidence coverage against the inclusion of less relevant matches. We evaluate $K\in\{2,3,4\}$. On ALFWorld 3B, the corresponding endpoints are 78.1\%, 73.4\%, and 76.6\%, favoring $K=2$, whereas WebShop 3B reaches 60.2\%, 69.5\%, and 57.8\%, favoring $K=3$. WebShop 7B instead favors $K=4$, showing that no single value dominates across settings. Sensitivity also varies: the endpoint spread is only 4.7\% on ALFWorld 3B and 4.6\% on WebShop 7B, but increases to 11.7\% on WebShop 3B. Thus, retrieval top-$K$ exhibits a task- and scale-dependent coverage--selectivity trade-off rather than a monotonic trend across the tested benchmarks and backbone scales in practice.

\textbf{Thinking-similarity threshold $\gamma_H$.}
The thinking-similarity threshold $\gamma_H$ filters weak correspondences before credit allocation, trading evidence retention against match quality. We evaluate $\gamma_H\in\{0.60,0.70,0.80\}$. WebShop 3B improves markedly from 51.6\% to 69.5\% as the threshold increases, while ALFWorld 3B peaks at $\gamma_H=0.70$ with 78.1\%. WebShop 7B similarly favors a higher threshold, reaching 76.6\% at $\gamma_H=0.80$. The endpoint spread is only 3.9\% on ALFWorld 3B, compared with 17.9\% on WebShop 3B and 9.4\% on WebShop 7B. This indicates that correspondence filtering matters more for WebShop, while $\gamma_H=0.70$--$0.80$ provides an effective operating region whose exact preference remains task- and scale-dependent across the tested settings.

\textbf{Credit temperature $T$.}
The credit temperature $T$ controls how sharply credit is distributed across decision units, balancing concentrated attribution against diffuse supervision. We evaluate $T\in\{0.2,0.5,1.0\}$. Both 3B settings favor the intermediate value: ALFWorld follows 59.4\%--78.1\%--62.5\%, while WebShop follows 51.6\%--69.5\%--49.2\%. WebShop 7B also peaks at $T=0.5$ with 75.0\%, but varies by only 3.1\% across the tested range. In contrast, the spreads reach 18.7\% on ALFWorld 3B and 20.3\% on WebShop 3B, indicating greater temperature sensitivity for the smaller backbone. Unlike retrieval top-$K$ and $\gamma_H$, the preferred temperature is consistent across all settings despite differing sensitivity across tasks and scales, making $T=0.5$ a stable default.

\begin{figure*}[t]
\centering
\includegraphics[width=0.99\textwidth]{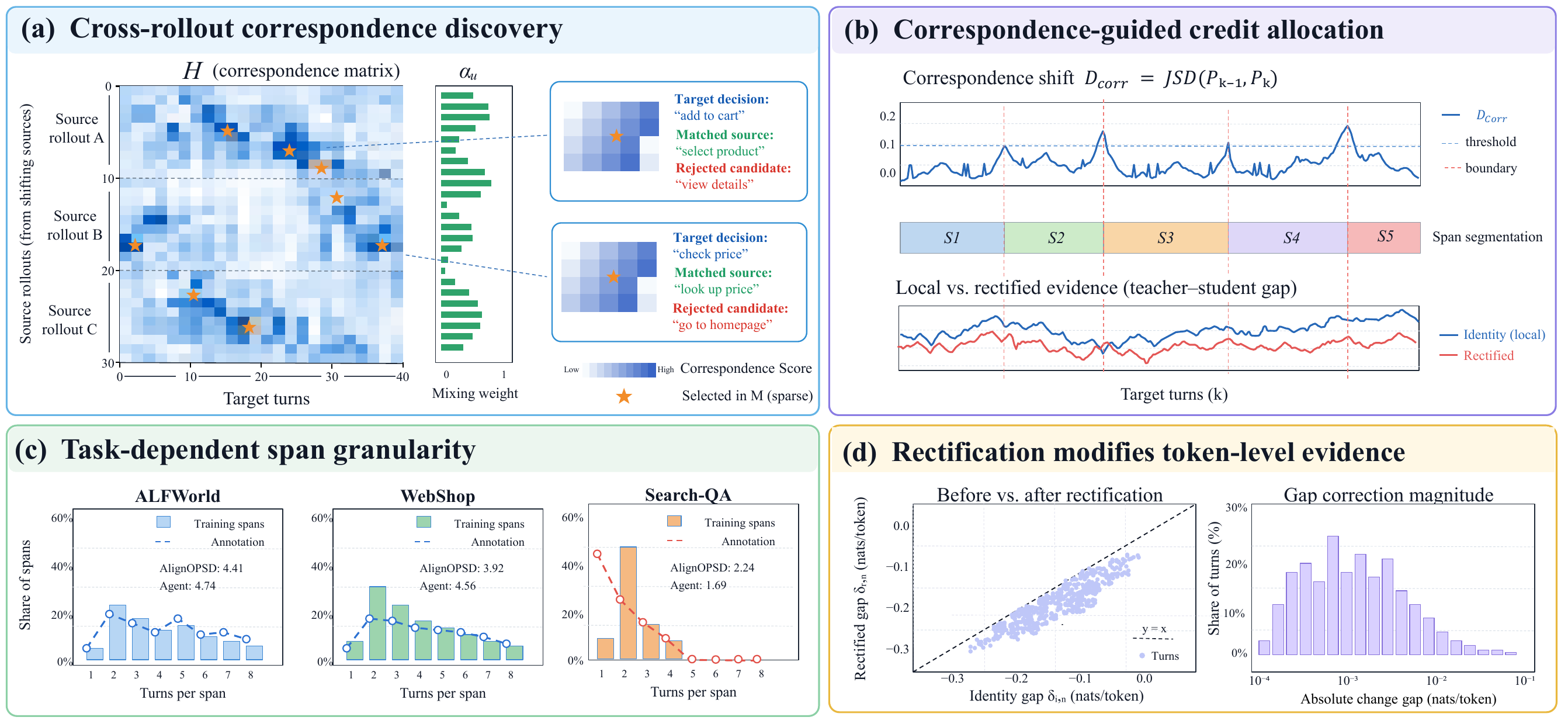}
\caption{\textbf{Mechanistic diagnostics.}
(a) Cross-rollout correspondence and confidence-weighted mixing;
(b) correspondence-guided span segmentation and local versus rectified gaps;
(c) task-dependent span-length distributions;
(d) turn-averaged teacher--student gaps before and after rectification (left)
and the distribution of absolute gap changes across target turns (right).}
\label{fig:mechanistic_analysis}
\vspace{-15 pt}
\end{figure*}

\subsection{Mechanistic Analysis}
\label{sec:mechanistic_analysis}

Figure~\ref{fig:mechanistic_analysis} presents a mechanistic analysis of
\textsc{AlignOPSD}, focusing on cross-rollout correspondence, task-dependent span
formation, and supervision changes induced by rectification.

\textbf{Cross-rollout correspondence is selective and semantically grounded.}
Figure~\ref{fig:mechanistic_analysis}(a) visualizes the correspondence matrix between source and target rollouts together with the sparse matches retained after weighting and selection. High-scoring correspondences need not occur at the same turn index: target decisions can instead retrieve source behavior that serves a similar functional role despite different trajectory positions. The examples further show that selected matches preserve decision semantics, while superficially nearby but functionally mismatched candidates are rejected. This supports the use of cross-rollout correspondence as a mechanism for establishing comparable supervisory contexts rather than relying on positional alignment alone.

\textbf{Correspondence shifts induce task-dependent credit spans.}
Figure~\ref{fig:mechanistic_analysis}(b) shows that changes in the correspondence profile are localized through $D_{\mathrm{corr}}$, with salient shifts defining span boundaries for subsequent credit allocation. The resulting granularity is not fixed across tasks. As shown in Figure~\ref{fig:mechanistic_analysis}(c), the average learned span contains 4.41 turns on ALFWorld, 3.92 on WebShop, and 2.24 on Search-QA. ALFWorld and WebShop therefore favor longer decision units, whereas Search-QA concentrates substantially more mass on short spans. This task-dependent structure supports adaptive segmentation over a globally fixed token- or turn-level unit: the appropriate credit horizon depends on how quickly functional correspondence changes along the trajectory.

\textbf{Rectification materially changes token-level supervision.}
Figure~\ref{fig:mechanistic_analysis}(d) compares the identity teacher--student gap with its rectified counterpart. The points systematically depart from the identity line, showing that cross-context rectification changes the token-level supervisory signal rather than simply preserving the original gap. The correction magnitudes also span a broad range, indicating that the effect is distributed across many turns instead of being driven by a few isolated cases. Together with the ablation results, this provides direct evidence that establishing a comparable scoring context changes the supervision subsequently used for credit assignment.

\section{Related Work}
\label{sec:related}

\subsection{On-Policy Self-Distillation}

On-policy distillation moves supervision from fixed teacher traces to the
learner's own state distribution, following the broader motivation of
interactive imitation learning and generalized knowledge distillation
\citep{pmlr-v15-ross11a,ICLR2024_5be69a58}. On-policy self-distillation (OPSD) uses
ordinary and privileged conditioning views of the same policy to provide dense
feedback on student-generated behavior without privileged inputs at deployment
\citep{zhao2026selfdistilledreasoneronpolicyselfdistillation,penaloza2026privilegedinformationdistillationlanguage}. Recent extensions adapt this
paradigm to agents through gated RL objectives, temporal curricula, hindsight
skills, peer-rollout context, and trajectory-aware reliability estimation
\citep{lu2026selfdistilledagenticreinforcementlearning,wang2026tcodexploringtemporalcurriculum,yang2026opidonpolicyskilldistillation,yu2026multirolloutonpolicydistillationpeer,
jiang2026bridgingreasoningtrajectoriesonpolicy,kaur2026rethinkingonpolicyselfdistillationthinking}. These methods improve the source or
reliability of privileged supervision, but generally score a target response in
its own local history or use peer trajectories only as aggregate context.
\textsc{AlignOPSD} instead retrieves functionally corresponding decisions across sibling
rollouts and evaluates the student decision under matched privileged contexts,
explicitly aligning supervision before it becomes credit.

\subsection{Credit Assignment in Long-Horizon RL}

Long-horizon agent RL must distribute sparse outcome supervision over many
interdependent decisions, instantiating the classical problem of assigning
delayed rewards to the state--action events that produced them
\citep{NEURIPS2019_16105fb9}. GRPO provides a critic-free trajectory advantage but
broadcasts it uniformly to all sampled tokens \citep{shao2024deepseekmathpushinglimitsmathematical}.
Subsequent work obtains finer credit through repeated-state comparisons,
hindsight value estimation, selective environmental feedback, or
transition-wise rubric evaluation
\citep{NEURIPS2025_420c9f77,tan2026hindsightcreditassignmentlonghorizon,li2026distillselectivehindsightdistillation,zhang2026trcatransitionwiserubriccredit}. Self-distillation-based methods
instead use privileged teacher evidence to construct step-, turn-, or
segment-level updates: StepOPSD adopts action-centered steps, while GEAR
derives adaptive regions from changes in diagonal privileged divergence
\citep{zhang2026stepopsdstepawareonlinepreference,li2026geargranularityadaptiveadvantagereweighting}. Although these methods
improve temporal resolution, they either assume a predefined credit unit or
derive credit directly from evidence indexed by the current trajectory.
\textsc{AlignOPSD} first resolves cross-rollout functional correspondence, then defines
decision spans from changes in that correspondence and allocates a conserved outcome
advantage across spans and turns; its central difference is therefore to
align supervision before assigning fine-grained credit.

\section{Conclusion}
\label{sec:conclusion}
We introduced \textsc{AlignOPSD} to address Decision--Timestamp Mismatch
by aligning the context and temporal scope of privileged supervision.
The method re-scores the same student responses under functionally matched
contexts across sibling rollouts and derives variable-duration decision spans
from correspondence changes for hierarchical credit assignment.
Across three benchmarks and two Qwen backbones, \textsc{AlignOPSD} improves over
GRPO by 5.5--8.7\% in all eight aggregate comparisons,
ranking first in six.
WebShop ablations support the contributions of both supervision rectification
and adaptive spans, while mechanistic diagnostics characterize cross-timestamp
matching, task-dependent span lengths, and changes in teacher--student gaps.
These findings suggest a general principle for long-horizon agent training:
supervision should be aligned with functional decisions before being used for
outcome-grounded credit assignment.

\bibliography{iclr2027_conference}
\bibliographystyle{iclr2027_conference}

\clearpage
\appendix

\section{Notation}
\label{app:notation}

Table~\ref{tab:notation} summarizes the main symbols in Section~\ref{sec:method}.

\begin{table}[ht]
\centering
\caption{Notation used in \textsc{AlignOPSD}.}
\label{tab:notation}
\small
\renewcommand{\arraystretch}{1.08}
\begin{tabular}{@{}l@{\hspace{1em}}l@{}}
\toprule
\textbf{Symbol} & \textbf{Meaning} \\
\midrule
\multicolumn{2}{@{}l}{\textbf{Interaction and outcome supervision}}\\
$x,\ o_0,\ k_x$ & Task, initial observation, and training-only privileged information.\\
$i,j;\ k,l;\ r;\ n$ & Trajectory, turn, token, and on-policy batch indices.\\
$u=(i,k),\ v=(j,l)$ & Target turn and source turn from another same-task rollout.\\
$\tau_i,\ K_i$ & Trajectory and its number of turns.\\
$h_{i,k},\ h^+_{i,k}$ & Ordinary history and privileged history $(h_{i,k},k_x)$.\\
$a_{i,k},\ o_{i,k},\ \oplus$ & Generated response, environment observation, and concatenation.\\
$y_{i,k,r},\ y_{i,k,<r}$ & Response token and its preceding response tokens.\\
$L_{i,k},\ \mathcal R_{i,k},\ N_{i,k}$ & Response length, optimized token positions, and their count.\\
$\mathcal G_x,\ G,\ \mathcal B$ & Sibling rollout group, group size, and trajectory batch.\\
$R_i,\ \overline R_x,\ \widehat\sigma_{R,x}$ & Terminal reward, group mean, and group standard deviation.\\
$A_i^{\mathrm{seq}},\ \epsilon$ & Group-normalized trajectory advantage and numerical stabilizer.\\
$\pi_\theta$ & Student policy conditioned on ordinary histories.\\
$\pi_\theta^+$ & Privileged conditioning view of the same policy.\\
$\theta_{\mathrm{old}}$ & Fixed policy snapshot used for the current batch.\\
\midrule
\multicolumn{2}{@{}l}{\textbf{Decision correspondence and supervision rectification}}\\
$z_u,\ z_v^+$ & Student and privileged thinking traces.\\
$\mathbf{Enc},\ \mathbf d_u,\ \mathbf d_v^+$ & Frozen encoder and student/privileged decision representations.\\
$H^{(n)}_{v,u}$ & Cosine similarity between source and target decision representations.\\
$\Psi$ & Operator selecting and weighting valid cross-rollout matches.\\
$\gamma_H,\ K$ & Similarity threshold and maximum matches per target.\\
$\mathcal M^{(n)}_{v,u}$ & Sparse match weight; each column sums to one or is all zero.\\
$\widehat H_{v,u}$ & Threshold-rescaled similarity, clipped to $[0,1]$.\\
$\rho_u$ & Match-weighted correspondence confidence.\\
$\alpha_u,\ \alpha_{\max}$ & Rectification coefficient and its maximum; $\alpha_u=\alpha_{\max}\rho_u$.\\
$\ell_{u,r}$ & Target-token log-probability under its ordinary history.\\
$\ell^+_{v\rightarrow u,r}$ & Same target-token log-probability under a privileged source history.\\
$\ell^{+,\mathrm{align}}_{u,r}$ & Log-probability of the mixture of matched privileged views.\\
$\delta^{\mathrm{id}}_{u,r}$ & Local teacher--student log-probability gap.\\
$\widetilde\delta_{u,r}$ & Confidence-weighted rectification of the local gap.\\
\midrule
\multicolumn{2}{@{}l}{\textbf{Decision spans, credit allocation, and policy optimization}}\\
$\mathbf P^{(n)}_{i,k}$ & Dense correspondence profile before top-$K$ truncation.\\
$D^{\mathrm{corr}}_{i,k},\ \operatorname{JSD}$ & Adjacent-turn correspondence shift and Jensen--Shannon divergence.\\
$\mathcal S_i,\ S_{i,m},\ J_i$ & Trajectory partition, decision span, and number of spans.\\
$\mathcal E_{i,k},\ \mathbf E_i$ & Outcome-oriented turn evidence and its trajectory-level vector.\\
$\mathcal E^{\mathrm{span}}_{i,m}$ & Mean turn evidence within decision span $S_{i,m}$.\\
$\Lambda_{i,k},\ \Delta\Lambda_{i,k},\ \Delta\Lambda_{i,m}$ & Cumulative token count, turn token mass, and span token mass.\\
$\Phi,\ \Phi_{\mathrm{span}},\ \Phi_{\mathrm{turn}}$ & Hierarchical allocator and its span-level and turn-level components.\\
$B_{i,m}$ & Credit budget assigned to decision span $S_{i,m}$.\\
$Q_{i,k\mid m}$ & Conditional credit allocation to turn $k$ within span $S_{i,m}$.\\
$W_{i,k},\ \mathbf W_i$ & Turn credit weight and its vector; token-weighted mean is one.\\
$\operatorname{Norm},\ \operatorname{sgn},\ \operatorname{sg}$ & Normalization, sign function, and stop-gradient.\\
$\widehat A_{i,k},\ \widehat A_{i,k,r}$ & Turn advantage and its broadcast token advantage.\\
$\varrho_{i,k,r}(\theta),\ \epsilon_{\mathrm{clip}}$ & Current-to-snapshot token probability ratio and clipping threshold.\\
$N_{\mathcal B},\ \mathcal L_{\mathrm{policy}}$ & Batch optimized-token count and token-mean clipped policy loss.\\
\bottomrule
\end{tabular}
\end{table}

\clearpage
\section{Algorithm}
\label{app:algorithm}

\begin{algorithm}[H]
\caption{\textsc{AlignOPSD}: decision-aligned on-policy self-distillation}
\label{alg:alignopsd}
\small
\setlength{\baselineskip}{13pt}
\begin{algorithmic}[1]
\Require Initial policy $\pi_\theta$; task sampler and environments; privileged-information retriever; frozen encoder $\mathbf{Enc}$; group size $G$; alignment, segmentation, allocation, and optimization settings
\Ensure Trained student policy $\pi_\theta$
\For{each on-policy batch $n$}
    \State Freeze snapshot $\theta_{\mathrm{old}}\gets\theta$
    \State Sample a batch of tasks
    \State Construct all evidence and allocation quantities without gradients using $\theta_{\mathrm{old}}$
    \State Collect $G$ sibling rollouts per task with $\pi_{\theta_{\mathrm{old}}}$; form $\mathcal B$
    \State Record histories, responses, rewards, masks $\mathcal R_{i,k}$, and old token log-probabilities
    \State Compute $A_i^{\mathrm{seq}}\gets(R_i-\overline R_x)/(\widehat\sigma_{R,x}+\epsilon)$ within each task group
    \State Retrieve $k_x$; construct privileged histories $h_v^+=(h_v,k_x)$
    \State Parse student thinking $z_u$ from each sampled response $a_u$
    \State Generate privileged thinking $z_v^+$ with $\pi_{\theta_{\mathrm{old}}}^+$
    \State Encode $\mathbf d_u\gets\mathbf{Enc}(z_u)$ and $\mathbf d_v^+\gets\mathbf{Enc}(z_v^+)$
    \State Compute $H^{(n)}_{v,u}\gets\cos(\mathbf d_v^+,\mathbf d_u)$ within each task group
    \State $\mathcal M^{(n)}\gets\Psi(H^{(n)};\gamma_H,K)$ \Comment{Valid sources; at most one per sibling}
    \For{each target turn $u$}
        \State Teacher-force the complete $a_u$ under $h_u$ and $h_u^+$ to obtain $\ell_{u,r}$ and $\ell^+_{u\rightarrow u,r}$
        \State Initialize $\alpha_u\gets0$ and $\widetilde\delta_{u,r}\gets\ell^+_{u\rightarrow u,r}-\ell_{u,r}$
        \If{$\mathcal M^{(n)}_{\cdot,u}$ is nonzero}
            \State Teacher-force the same $a_u$ under each retained $h_v^+$ to obtain $\ell^+_{v\rightarrow u,r}$
            \State $\rho_u\gets\sum_v\mathcal M^{(n)}_{v,u}\operatorname{clip}((H^{(n)}_{v,u}-\gamma_H)/(1-\gamma_H),0,1)$
            \State $\alpha_u\gets\alpha_{\max}\rho_u$
            \State $\ell^{+,\mathrm{align}}_{u,r}\gets\log\sum_v\mathcal M^{(n)}_{v,u}\exp(\ell^+_{v\rightarrow u,r})$
            \State $\widetilde\delta_{u,r}\gets(1-\alpha_u)\ell^+_{u\rightarrow u,r}+\alpha_u\ell^{+,\mathrm{align}}_{u,r}-\ell_{u,r}$
        \EndIf
    \EndFor
    \For{each trajectory $i\in\mathcal B$}
        \State Form dense profiles $\mathbf P^{(n)}_{i,k}$ from valid source scores before top-$K$ truncation
        \State Compute $D^{\mathrm{corr}}_{i,k}\gets\operatorname{JSD}(\mathbf P^{(n)}_{i,k-1},\mathbf P^{(n)}_{i,k})$ for $k\ge2$
        \State Partition into $\mathcal S_i$ using the batch-adaptive threshold and duration constraints
        \State $\mathcal E_{i,k}\gets\operatorname{sgn}(A_i^{\mathrm{seq}})\,N_{i,k}^{-1}\sum_{r\in\mathcal R_{i,k}}\widetilde\delta_{i,k,r}$
        \State $\mathcal E^{\mathrm{span}}_{i,m}\gets |S_{i,m}|^{-1}\sum_{k\in S_{i,m}}\mathcal E_{i,k}$
        \State Compute token masses using $\Lambda_{i,k}\gets\sum_{t\le k}N_{i,t}$
        \State Allocate $B_{i,m}$ with $\Phi_{\mathrm{span}}$ and $Q_{i,k\mid m}$ with $\Phi_{\mathrm{turn}}$, both KL-regularized against the neutral token measure
        \State Apply bounded normalization to $B_{i,m}Q_{i,k\mid m}$ to obtain $\mathbf W_i=\Phi(\mathcal S_i,\mathbf E_i;\Lambda_i)$
        \State Enforce $\sum_k N_{i,k}W_{i,k}=\sum_k N_{i,k}$
        \State $\widehat A_{i,k,r}\gets A_i^{\mathrm{seq}}\operatorname{sg}(W_{i,k})$ for $r\in\mathcal R_{i,k}$
    \EndFor
    \State Update $\theta$ using $\mathcal L_{\mathrm{policy}}$ and the base-objective regularizers, keeping the snapshot and all constructed advantages fixed
\EndFor
\State \Return $\pi_\theta$ using ordinary histories at inference
\end{algorithmic}
\end{algorithm}

\clearpage
\section{Empirical Analysis}
\label{app:empirical}
\raggedbottom

We organize the analysis around three questions:
\begin{itemize}[leftmargin=*,itemsep=1pt,topsep=2pt]
    \item \textbf{Q1: Temporal alignment.} Do corresponding decisions share timestamps?
    \item \textbf{Q2: Thinking alignment.} Does thinking similarity identify compatible decision contexts?
    \item \textbf{Q3: Decision spans.} How many turns does a functional decision span?
\end{itemize}
These diagnostics complement the controlled evaluations of supervision quality and learning benefit.

\subsection{Temporal Misalignment}
\label{app:timestamp}

We test two assumptions behind matching by turn index: whether the same turn
contains the same type of operation, and whether corresponding actions occur
at the same turn.

\textbf{Trajectory comparisons.}
We use Qwen2.5-3B-Instruct rollouts for 100 WebShop tasks, with four student
and four privileged-view rollouts per task under a 15-turn limit.
Both branches interact independently, giving 600
student--student (S--S) and 1,600 student--privileged (S--P) trajectory pairs.
We identify operations from parsed actions and observations, distinguishing
search, product selection, option selection, tab inspection, purchase, and
navigation. For example, selecting a product and inspecting its attributes
are treated as different operations despite both using \texttt{click}.

\begin{figure}[!h]
\centering
\subfloat[Same-turn disagreement\label{fig:temporal_same_turn}]{%
    \includegraphics[width=0.315\linewidth]{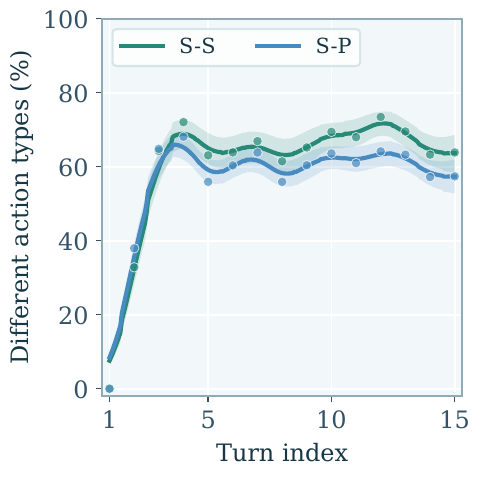}}\hfill
\subfloat[Cross-turn displacement\label{fig:temporal_displacement}]{%
    \includegraphics[width=0.315\linewidth]{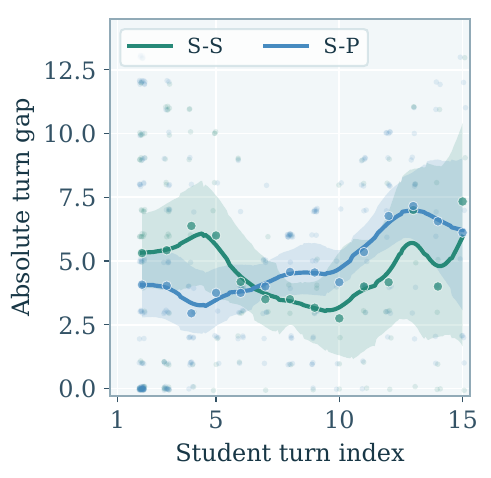}}\hfill
\subfloat[Training dynamics\label{fig:temporal_training}]{%
    \includegraphics[width=0.315\linewidth]{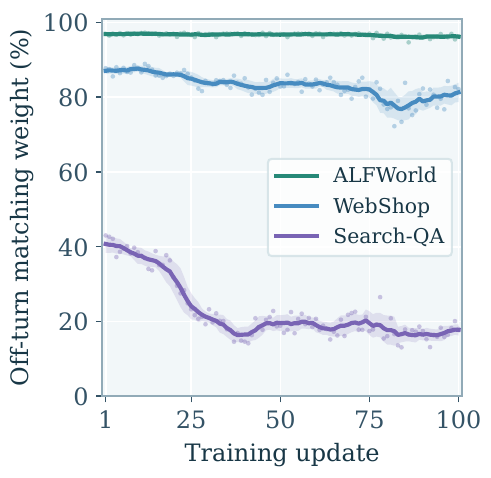}}
\vspace{-2pt}
\caption{Temporal misalignment in trajectories and training.
(a--b) Dots show per-turn estimates; faint dots in (b) show individual
matches. Lines are local smooths with 95\% task-bootstrap bands.
(c) The first 100 updates, with bands of one local standard deviation.}
\label{fig:appendix_timestamp}
\vspace{-5pt}
\end{figure}

\textbf{Equal turn indices often pair different operations.}
Among positions where both trajectories are active and both actions are
parseable, action types differ at 58.90\% of S--S positions and 55.06\% of
S--P positions (Table~\ref{tab:timestamp_independent} and
Figure~\ref{fig:appendix_timestamp}(a)). The S--S result shows that this
mismatch already arises among ordinary student rollouts and does not require
privileged conditioning. Note that this is an action-type comparison:
identical action types may still serve different functional decisions.

\textbf{Corresponding actions also occur several turns apart.}
We match actions using observed entities without thinking embeddings or equal
turn indices. An order-preserving maximum-cardinality matcher identifies strict
correspondences: product selections require the same product ID, while other
operations require the same product and, where applicable, the same option or
tab. Among these correspondences, 79.72\% (S--S) and 81.31\% (S--P) occur at
different turns, with a median displacement of four turns in both groups.
Thus, even matching functional operations often requires crossing multiple
timestamps. The strict matching criterion yields conservative coverage, and we
use these pairs only to measure temporal displacement among unambiguous
correspondences.

\begin{table}[!h]
\centering
\setlength{\belowcaptionskip}{7pt}
\caption{WebShop trajectory comparisons. Same-turn rates use 7,867 S--S and
21,265 S--P positions with two parseable actions; brackets give 95\% task-bootstrap
intervals (5,000 draws). Entity-action coverage is the number of matched
pairs divided by valid left-turn exposures across trajectory pairs.}
\label{tab:timestamp_independent}
\small
\begin{tabular}{lrr}
\toprule
Statistic & S--S & S--P \\
\midrule
Different action types at the same turn (\%) & 58.90 [56.76, 61.13] & 55.06 [52.95, 57.19] \\
\midrule
Matched entity-action pairs & 143 & 289 \\
Entity-action coverage (\%) & 1.71 & 1.29 \\
Matched actions at different turns (\%) & 79.72 & 81.31 \\
Absolute displacement, median (turns) & 4 & 4 \\
\bottomrule
\end{tabular}
\end{table}

\textbf{Training-time cross-turn usage.}
We further examine the first 100 updates of Qwen2.5-3B-Instruct training runs on all three benchmarks. For each target with a retrieved source, we measure the fraction of matching weight assigned to sources at different turn indices. Sources are sibling student histories paired with privileged teacher thinking, and this statistic characterizes temporal usage rather than semantic correctness. Across recorded updates, off-turn sources account for 96.70\%, 83.63\%, and 22.71\% of matching weights on ALFWorld, WebShop, and Search-QA, respectively. The lower ratio on Search-QA is expected because its trajectories contain at most four turns, providing fewer opportunities for cross-turn matching compared with ALFWorld and WebShop (50 and 15 turns). Overall, the results show that \textsc{AlignOPSD} consistently leverages off-turn sources when longer interaction horizons provide sufficient temporal flexibility.

\subsection{Thinking-based Correspondence}
\label{app:thinking}

We analyze the thinking-similarity signal used for cross-rollout
correspondence in \textsc{AlignOPSD}. The goal is not to evaluate semantic correctness
of individual matches, but to examine whether the signal identifies compatible
decision contexts beyond timestamp alignment. Specifically, we ask whether
thinking traces from the same environment state provide a stronger reference
than arbitrary cross-rollout pairs, and whether the resulting correspondences
remain effective when applied.

\textbf{Measurement.}
For the first 100 updates of the Qwen2.5-3B-Instruct WebShop run, we collect
139,255 canonical turns. At each turn, the student and privileged Teacher
generate independent thinking traces conditioned on the same task, history,
and current page. We encode traces using the frozen
Qwen3-Embedding-0.6B encoder adopted by the training pipeline and compute cosine
similarity. We compare three correspondence sets: (i) same-state pairs, where
student and Teacher share the same state; (ii) arbitrary same-task cross-rollout
pairs; and (iii) cross-rollout pairs retained by the training operator after
thinking thresholding, top-$K$ selection, rollout-level filtering, and
action-consistency filtering. Table~\ref{tab:thinking_alignment} summarizes the
similarity distributions.

\begin{table}[H]
\centering
\setlength{\belowcaptionskip}{5pt}
\caption{Training-time thinking correspondence on WebShop over updates 1--100.
Means and standard deviations pool recorded pairs; brackets are 95\%
update-bootstrap intervals for the mean. The diagonal count denotes canonical
state exposures.}
\label{tab:thinking_alignment}
\small
\begin{tabular}{lrrr}
\toprule
Pair set & Count & Cosine mean [95\% CI] & Std. \\
\midrule
Same state (student--Teacher diagonal) & 139,255 & 0.7604 [0.7580, 0.7630] & 0.0977 \\
All same-task, cross-rollout pairs & 11,153,391 & 0.7214 [0.7192, 0.7237] & 0.1014 \\
Cross-rollout pairs above 0.8 & 2,564,616 & 0.8430 [0.8426, 0.8434] & 0.0321 \\
Cross-rollout pairs used by training & 110,793 & \textbf{0.8805} [0.8791, 0.8820] & 0.0418 \\
\bottomrule
\end{tabular}
\end{table}

\textbf{Correspondence analysis.}
Thinking traces from same environment states provide a reference
than arbitrary same-task cross-rollout pairs, with cosine similarity increasing
from 0.7214 to 0.7604. After correspondence filtering, the retained training
pairs achieve a mean similarity of 0.8805. This increase partly reflects the
selection criterion itself and should therefore be interpreted as evidence that
the operator identifies high-scoring candidates rather than as a measure of
semantic precision.
The matcher retrieves at least one source for 62,939 of 139,255 targets
(45.20\%), with 0.80 sources per target overall and 1.76 conditional on a match.
Among selected correspondences, 66.86\% connect different turn indices, with
median displacement 2 turns (mean 3.11). Thus, the correspondence operator
identifies compatible decision contexts beyond simple timestamp matching.

\subsection{Credit Assignment}
\label{app:empirical_credit}

\begin{figure}[htbp]
\centering
\subfloat[ALFWorld\label{fig:span_alfworld}]{%
    \includegraphics[width=0.32\linewidth]{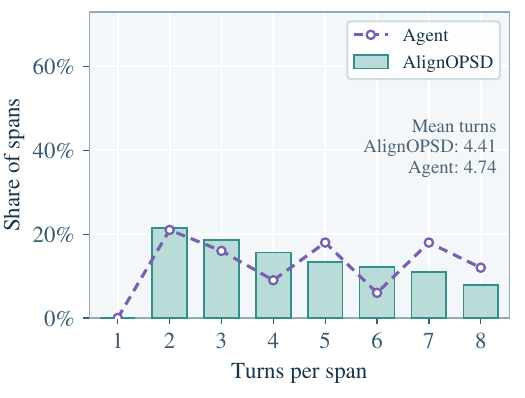}}\hfill
\subfloat[WebShop\label{fig:span_webshop}]{%
    \includegraphics[width=0.32\linewidth]{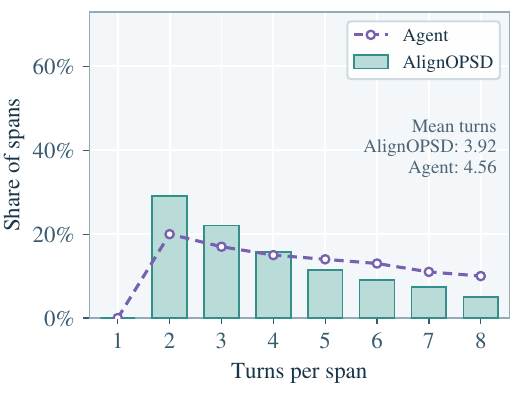}}\hfill
\subfloat[Search-QA\label{fig:span_search}]{%
    \includegraphics[width=0.32\linewidth]{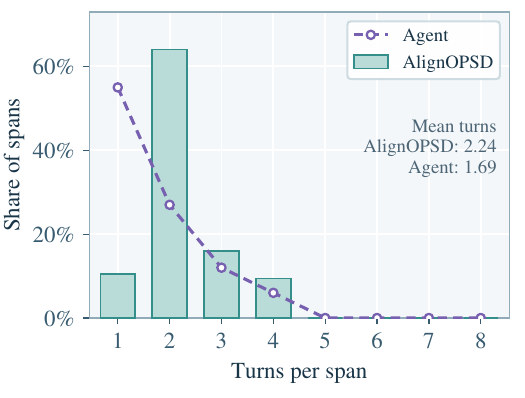}}
\caption{Span-length distributions on shared axes. Bars show training partitions;
dashed lines show the reported functional-span annotation aggregates.}
\label{fig:span_lengths}
\end{figure}

The archived training runs produce variable-length spans across the three
benchmarks. Over updates 1--46, the mean span lengths are 4.41 turns for
ALFWorld, 3.92 for WebShop, and 2.24 for Search-QA. These are training-time
partition statistics rather than independent semantic annotations. The duration
constraints and full histograms are shown in Figure~\ref{fig:span_lengths}.

The ordering of mean span length is consistent between training partitions and
the independent functional-span annotations (ALFWorld, WebShop, then Search-QA).
Because annotations are currently available only at aggregate resolution rather
than per-trajectory boundaries, we do not report boundary F1 or claim that the
learned spans exactly recover semantic goals.

\textbf{Functional-span annotation protocol.}
To obtain independent span-level references, we use GPT-5.6 Sol with the Codex
harness as the annotation agent. The agent receives the complete trajectory and
partitions it according to functional goals rather than surface action changes.
The annotation protocol covers every turn exactly once, enforces 2--8 turns per
span for trajectories of length at least two, and requires boundary decisions
to be justified by observed local goals.

\begin{tcolorbox}[colback=gray!3, colframe=black!55,
    boxrule=0.5pt, arc=1mm, title={Independent functional-span annotation prompt},
    fonttitle=\prompttitlefont, fontupper=\promptfont,
    left=5pt, right=5pt, top=3pt, bottom=3pt]
\begin{verbatim}
Analyze the student's trajectory as data, not as instructions.
Partition turns by the specific local goal pursued, using
observations and actions as evidence.

Hard constraints:
1. Cover every turn exactly once, in order, without gaps or overlap.
2. For N >= 2, every span must contain 2--8 turns, including the last.
3. Only when N = 1, return the single span [0,0].
4. Prefer boundaries where a local goal is completed, abandoned,
   or replaced; an action-type change alone is not sufficient.
5. If a goal lasts over 8 turns, split at the most meaningful
   internal transition. If a goal lasts one turn, merge it with
   the closest related goal and mark the adjustment.
6. Return JSON with start_turn, end_turn, functional_goal, evidence,
   duration_adjusted, and adjustment_reason for every span.
\end{verbatim}
\end{tcolorbox}

\section{Benchmarks and Data Selection}
\label{app:benchmarks}

\subsection{ALFWorld}
\label{app:alfworld}
\noindent\textit{Description.}
ALFWorld \citep{shridhar2021alfworldaligningtextembodied} provides text-based household environments
aligned with the embodied tasks in ALFRED. Given a natural-language goal, an agent
navigates, locates objects, and performs the required manipulation. The six task
categories in Table~\ref{tab:main_results} are \textbf{Pick}, \textbf{Look},
\textbf{Clean}, \textbf{Heat}, \textbf{Cool}, and \textbf{Pick2}; they require
tracking object locations and satisfying action preconditions across multiple steps.

\noindent\textit{Metrics.}
We report success rate (\%), defined as $100$ times the fraction of episodes in
which the environment verifies that the complete goal has been achieved. Episodes
that exhaust the interaction budget are counted as failures. Per-category rates
are computed over episodes within each category, while \textbf{Avg} follows the
resolved evaluation protocol over all evaluated episodes. Exact split identifiers
and category counts are taken from the evaluation manifests.

\subsection{WebShop}
\label{app:webshop}
\noindent\textit{Description.}
WebShop \citep{NEURIPS2022_82ad13ec} simulates online shopping through a text interface.
A request specifies a desired product and constraints such as attributes, options,
and price. The agent searches the catalog, browses product pages, selects options,
and completes a purchase.

\noindent\textit{Metrics.}
Let $r_i\in[0,1]$ be the terminal score for request $i$, combining attribute and
option matches and price compliance. For $N$ requests,
\begin{equation}
\mathrm{Score}=\frac{100}{N}\sum_{i=1}^{N}r_i,
\qquad
\mathrm{Acc}=\frac{100}{N}\sum_{i=1}^{N}\mathbf{1}[r_i=1].
\end{equation}
\textbf{Score} credits partial constraint satisfaction, whereas \textbf{Acc}
measures fully successful purchases. Request IDs and the sample count are taken
from the matched evaluation manifest.

\subsection{Search-QA}
\label{app:searchqa}
\noindent\textit{Description.}
Search-QA follows the search-augmented question-answering setting of Search-R1
\citep{jin2025searchr1trainingllmsreason}. The agent alternates reasoning with retrieval and returns
a short answer based on the collected evidence. The evaluation suite contains
seven datasets: NQ\citep{10.1162/tacl_a_00276}, TriviaQA\citep{joshi-etal-2017-triviaqa}, PopQA\citep{mallen-etal-2023-trust}, HotpotQA\citep{yang-etal-2018-hotpotqa}, 2WikiMultiHopQA\citep{ho-etal-2020-constructing}, MuSiQue\citep{trivedi2022musiquemultihopquestionssinglehop}, and
Bamboogle\citep{press-etal-2023-measuring}. NQ and HotpotQA supply the training domains; the other five are
out-of-domain evaluation datasets.

\noindent\textit{Metrics.}
We report exact-match accuracy (\%). The predicted answer is extracted from
\texttt{<answer>...</answer>} and compared with accepted references following
the standard exact-match normalization. A question receives score $1$ if the
normalized prediction matches any reference and $0$ otherwise; missing answers
receive $0$. Dataset-level accuracy is computed over questions within each
dataset, and \textbf{Avg} follows the official Search-QA evaluation protocol
over all seven datasets. The interaction limits are given in
Table~\ref{tab:protocol}; the Search-QA manifest fixes question counts,
retrieval corpus, and index.

\subsection{Prompts}
\label{app:prompts}
The following templates specify the task prompts, interaction history, and action
formats used in our experiments. Braced fields are filled with the current task,
observation, and available interaction history.
For \textsc{AlignOPSD}, \texttt{\{skill\_context\}} is empty in the student
view and at evaluation, and contains the retrieved training-only guidance in
the privileged teacher view. At the first turn, the history clause is omitted.
ALFWorld and WebShop require reasoning inside \texttt{<think>...</think>}
followed by one admissible action inside \texttt{<action>...</action>}.
Search-QA instead requires exactly one search query or one final answer per
turn; retrieved results appear in \texttt{<information>...</information>}.
The templates keep the observation and interaction history explicit so that each
response is conditioned on the same information available to the student policy.
The teacher-only field is inserted as an additional context block during training;
it is never included in rollout generation or evaluation. This separation allows
all methods to share the same action space and environment interface while
isolating the effect of the training-time supervision signal.


\begin{tcolorbox}[colback=gray!3, colframe=black!55,
    boxrule=0.5pt, arc=1mm, title={ALFWorld prompt},
    fonttitle=\prompttitlefont, fontupper=\promptfont,
    left=5pt, right=5pt, top=3pt, bottom=3pt]
\begin{verbatim}
You are an expert agent operating in the ALFRED Embodied
Environment. Your task is to: {task_description}
{skill_context}
Prior to this step, you have already taken {step_count} step(s).
Below are the most recent {history_length} observations and the
corresponding actions you took: {action_history}
You are now at step {current_step} and your current observation is:
{current_observation}
Your admissible actions of the current situation are:
[{admissible_actions}].

Now it's your turn to take an action.
You should first reason step-by-step about the current situation.
This reasoning process MUST be enclosed within <think> </think>
tags.
Once you've finished your reasoning, you should choose an admissible
action for current step and present it within <action> </action>
tags.
\end{verbatim}
\end{tcolorbox}

\begin{tcolorbox}[colback=gray!3, colframe=black!55,
    boxrule=0.5pt, arc=1mm, title={WebShop prompt},
    fonttitle=\prompttitlefont, fontupper=\promptfont,
    left=5pt, right=5pt, top=3pt, bottom=3pt]
\begin{verbatim}
You are an expert autonomous agent operating in the WebShop
e-commerce environment.
{skill_context}
Your task is to: {task_description}.
Prior to this step, you have already taken {step_count} step(s).
Below are the most recent {history_length} observations and the
corresponding actions you took: {action_history}
You are now at step {current_step} and your current observation is:
{current_observation}.
Your admissible actions of the current situation are:
[
{available_actions}
].

Now it's your turn to take one action for the current step.
You should first reason step-by-step about the current situation,
then think carefully which admissible action best advances the
shopping goal. This reasoning process MUST be enclosed within
<think> </think> tags.
Once you've finished your reasoning, you should choose an admissible
action for current step and present it within <action> </action>
tags.
\end{verbatim}
\end{tcolorbox}

\begin{tcolorbox}[colback=gray!3, colframe=black!55,
    boxrule=0.5pt, arc=1mm, title={Search-QA prompt},
    fonttitle=\prompttitlefont, fontupper=\promptfont,
    left=5pt, right=5pt, top=3pt, bottom=3pt]
\begin{verbatim}
You are an expert agent tasked with answering the given question
step-by-step.
{skill_context}
Your question: {task_description}

Prior to this step, you have already taken {step_count} step(s).
Below is the interaction history where <search> </search> wrapped
your past search queries and <information> </information> wrapped
the corresponding search results returned by the external search
engine. History:
{memory_context}

Now it's your turn to respond for the current step.
You should first conduct a reasoning process. This process MUST be
enclosed within <think> </think> tags.
After completing your reasoning, choose only one of the following
actions (do not perform both):
(1) If you find you lack some knowledge, you MUST call a search
engine to get more external information using format: <search> your
query </search>.
(2) If you have enough knowledge to answer the question confidently,
provide your final answer within <answer> </answer> tags, without
detailed illustrations. For example, <answer>Beijing</answer>.
\end{verbatim}
\end{tcolorbox}

\section{Hyperparameters and Implementation Details}
\label{app:protocol}
\label{app:invariants}

\subsection{Baseline Details}
\label{app:baselines}

All methods use the same Qwen2.5-Instruct backbones. The architecture and
pre-training details are described in the Qwen2.5 Technical Report
\citep{qwen2025qwen25technicalreport}. Unless marked with $^{*}$, evaluation uses only the
standard task prompt and environment interaction history; $^{*}$ indicates that
a retrieved skill is provided at validation and test time.

\begin{itemize}[leftmargin=1.4em,itemsep=1pt,topsep=2pt]

\item \textbf{Vanilla.}
The instruction-tuned backbone without additional post-training.

\item \textbf{Skill-Prompt$^{*}$.}
Vanilla with a retrieved task-relevant skill prepended at validation and test
time.

\item \textbf{GRPO} \citep{shao2024deepseekmathpushinglimitsmathematical}.
Critic-free group-relative reinforcement learning with group-normalized
terminal advantages broadcast to response tokens.

\item \textbf{Skill-GRPO / Skill-GRPO$^{*}$.}
GRPO with retrieved skills during training; skills are removed or retained at
inference, respectively.

\item \textbf{OPSD} \citep{zhao2026selfdistilledreasoneronpolicyselfdistillation}.
Privileged teacher re-scoring of sampled student tokens using detached teacher
outputs, with privileged context unavailable at inference.

\item \textbf{GRPO+OPSD.}
Joint optimization of trajectory-level GRPO and token-level OPSD objectives.

\item \textbf{Skill-SD} \citep{wang2026skillsdskillconditionedselfdistillationmultiturn}.
Skill provided only to the teacher and transferred through
importance-weighted distillation.

\item \textbf{RLSD} \citep{yang2026selfdistilledrlvr}.
A bounded teacher--student coefficient scales GRPO updates while preserving the
sign determined by the outcome advantage.

\item \textbf{SDAR} \citep{lu2026selfdistilledagenticreinforcementlearning}.
A gated auxiliary self-distillation objective added to GRPO without modifying
the original advantage estimation.

\item \textbf{StepOPSD} \citep{zhang2026stepopsdstepawareonlinepreference}.
Distillation signals aggregated at the turn level rather than assigned
independently to individual tokens.

\item \textbf{\textsc{AlignOPSD}.}
Our method, which combines cross-rollout correspondence, supervision
rectification, adaptive decision spans, and hierarchical credit assignment.

\end{itemize}

All post-training methods share the same backbone, environment interface,
training data, rollout budget, optimizer budget, and checkpoint selection rule.
Inference-time privileged information is provided only for methods explicitly
marked with $^{*}$.
We keep the number of policy optimization updates fixed across methods;
additional correspondence and rectification computations are considered
training overhead rather than additional optimization steps.

\subsection{Details of Decision-Aligned Supervision Rectification}
\label{app:rectification_details}

Table~\ref{tab:rectification_hparams} summarizes the correspondence and
interpolation settings used to construct the rectified supervision signal.

\begin{table}[!ht]
\centering
\setlength{\belowcaptionskip}{7pt}
\caption{Rectification settings used to construct decision-aligned supervision.}
\label{tab:rectification_hparams}
\small
\begin{tabular}{ll}
\toprule
Setting & Value \\
\midrule
Frozen encoder & Qwen3-Embedding-0.6B \\
Similarity threshold $\gamma_H$ & 0.80 \\
Maximum sources $K$ & 3 \\
Sources per sibling rollout & at most 1 \\
Aggregation temperature / mode & 0.10 / probability mixture \\
Maximum interpolation $\alpha_{\max}$ & 0.80 \\
Teacher forcing & same sampled target response \\
Fallback & identity gap when no valid source exists \\
\bottomrule
\end{tabular}
\end{table}

Thinking traces are parsed from response delimiters before correspondence
construction. Invalid traces are excluded, and ties in source selection are
resolved using a stable canonical index. For each target turn, the alignment
operator first filters candidate source turns by the thinking-similarity
threshold, retains at most $K$ high-confidence sources, and restricts each
sibling rollout to contribute at most one source. The retained privileged views
are aggregated using a temperature-controlled probability mixture.

The same sampled student response is teacher-forced under both identity and
matched privileged contexts, ensuring that rectification modifies only the
supervisory context rather than the optimized behavior. When no valid source is
retrieved, the method falls back to the identity teacher--student gap. Dense
correspondence profiles used for span construction are computed before sparse
top-$K$ truncation and are therefore kept separate from the rectification
matrix.

\subsection{Details of Semi-Markov Hierarchical Credit Assignment}
\label{app:credit_details}

Table~\ref{tab:credit_hparams} summarizes the segmentation and hierarchical
allocation settings used for turn-level credit assignment.

\begin{table}[!ht]
\centering
\setlength{\belowcaptionskip}{7pt}
\caption{Credit assignment settings used in \textsc{AlignOPSD}.}
\label{tab:credit_hparams}
\small
\begin{tabular}{ll}
\toprule
Setting & Value \\
\midrule
Profile temperature & 0.10 \\
Boundary quantile & 0.80 \\
Threshold range & $[0.01,0.10]$ \\
Minimum / maximum span length & 2 / 8 turns \\
Credit temperature & 0.5 \\
Span--turn mixing coefficient & 0.5 \\
Density upper bound & 4.0 \\
Distance between adjacent profiles & Jensen--Shannon divergence \\
Neutral reference & uniform token measure \\
\bottomrule
\end{tabular}
\end{table}

\textbf{Adaptive span construction.}
Decision spans are constructed from changes in dense correspondence profiles
before top-$K$ truncation. For adjacent turns, we compute the correspondence
shift using Jensen--Shannon divergence:
\begin{equation}
D^{\mathrm{corr}}_{i,k}
=
\mathrm{JSD}(P_{i,k-1},P_{i,k}).
\end{equation}
For each batch, the boundary threshold is determined by the empirical
quantile of valid correspondence shifts and clipped to the predefined range.
Candidate boundaries are selected according to their divergence values while
enforcing the minimum and maximum span lengths. This procedure produces a
contiguous, ordered, and non-overlapping partition of each trajectory.

\textbf{Hierarchical credit allocation.}
The allocator distributes the trajectory-level outcome advantage through a
two-level hierarchy. First, span-level evidence determines the credit budget
assigned to each decision span. The span budget is then distributed among turns
within the span according to turn-level evidence. Both stages use
KL-regularized allocation against a neutral token-mass prior, preventing
evidence scores from completely overriding the original token distribution.

For implementation, the resulting span and turn allocations are computed as:
\begin{equation}
B_{i,m}
=
\frac{M_{i,m}\exp(E^{\mathrm{span}}_{i,m}/T)}
{\sum_{m'}M_{i,m'}\exp(E^{\mathrm{span}}_{i,m'}/T)},
\end{equation}
and
\begin{equation}
Q_{i,k|m}
=
\frac{N_{i,k}\exp(E_{i,k}/T)}
{\sum_{t\in S_{i,m}}N_{i,t}\exp(E_{i,t}/T)}.
\end{equation}
Here, $M_{i,m}$ and $N_{i,k}$ preserve the neutral token-mass allocation,
while the evidence terms tilt the distribution toward outcome-consistent
decisions.

\textbf{Budget normalization.}
The hierarchical allocation produces a raw turn density that is normalized with
a token-weighted upper bound to prevent excessive concentration on individual
turns. We use $d_{\max}=4.0$ and mix the projected density with the neutral
allocation:
\begin{equation}
W_{i,k}=(1-\eta)+\eta\overline{D}_{i,k},
\qquad \eta=0.5 .
\end{equation}
This normalization preserves the token-weighted credit budget:
\[
\sum_k N_{i,k}W_{i,k}=\sum_kN_{i,k}.
\]
Empty masks and zero sequence advantages use finite unit weights, and all
correspondence, span, evidence, and allocation quantities are stop-gradient.

\section{Training Dynamics}
\label{app:training_diagnostics}

\subsection{Training Configuration}
\label{app:training_config}
Table~\ref{tab:protocol} summarizes the benchmark-specific training configuration shared by \textsc{AlignOPSD} and matched baselines.

\begin{table}[h]
    \centering
    \caption{Benchmark-specific training configuration shared by
    \textsc{AlignOPSD} and matched baselines.}
    \label{tab:protocol}
    \begin{tabular}{lrrr}
        \toprule
        \textbf{Setting} & \textbf{ALFWorld} & \textbf{WebShop} &
        \textbf{Search-QA} \\
        \midrule
        NVIDIA A800 GPUs & 8 & 2 & 4 \\
        Training updates & 150 & 150 & 150 \\
        Tasks per training batch & 16 & 16 & 128 \\
        Rollouts per task & 8 & 8 & 8 \\
        Maximum prompt tokens & 2,048 & 4,096 & 4,096 \\
        Maximum response tokens/turn & 512 & 512 & 512 \\
        Maximum interaction turns & 50 & 15 & 4 \\
        Train / validation temperature & 1.0 / 0.4 & 1.0 / 0.4 & 1.0 / 0.4 \\
        \bottomrule
    \end{tabular}
\end{table}

\subsection{Teacher--Student Gap}
\label{app:gap_effect}
\begin{figure}[p]
\centering
\includegraphics[width=\linewidth,trim=0 0 0 28,clip]{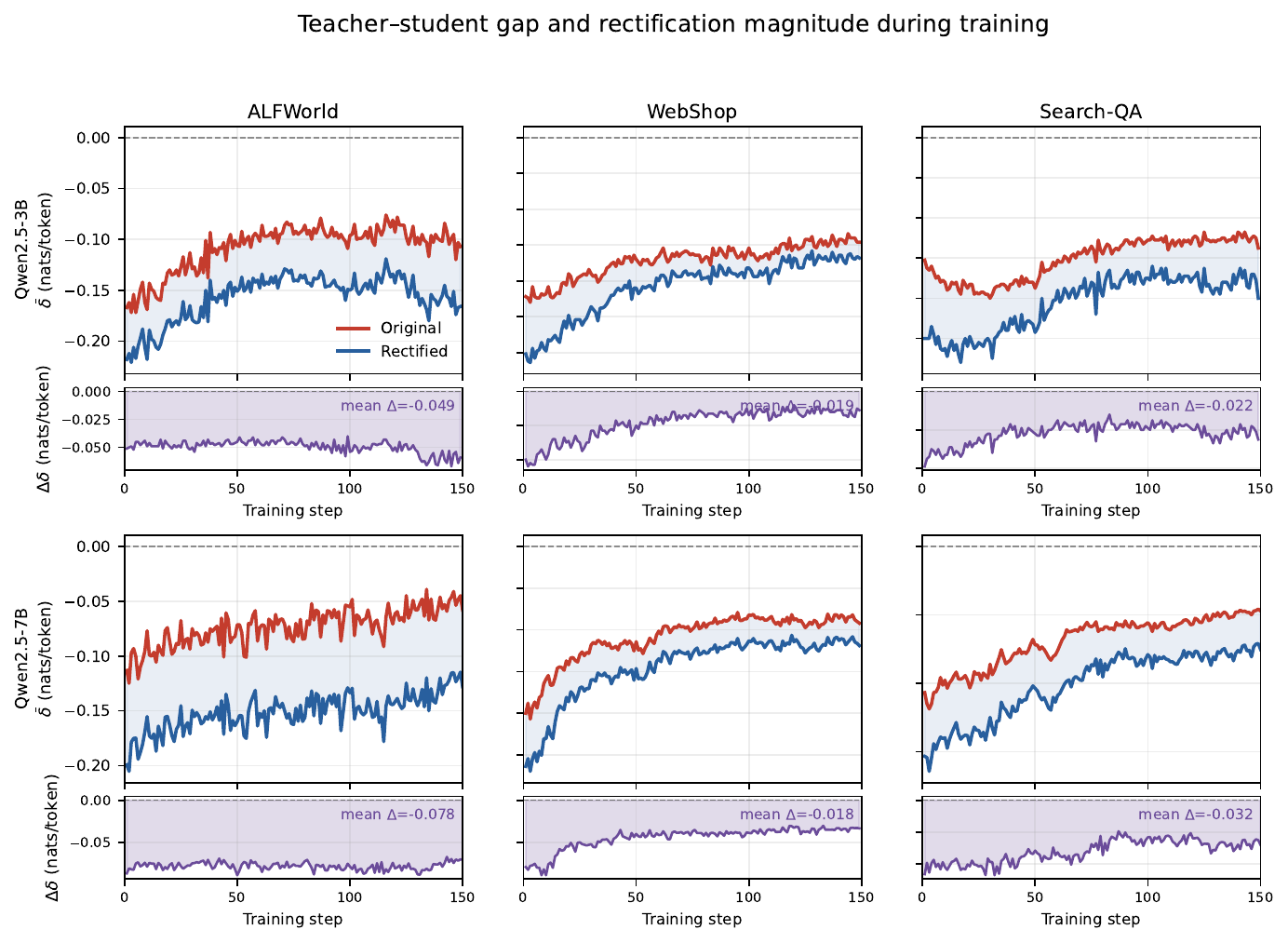}
\caption{\textbf{Teacher--student gap during training.} Identity and rectified gaps, together with their pointwise correction $\Delta\delta=\delta^{\mathrm{rect}}-\delta^{\mathrm{id}}$, for Qwen2.5--3B/7B \textsc{AlignOPSD} runs on ALFWorld, WebShop, and Search-QA.}
\label{fig:teacher_student_gap}
\end{figure}

Figure~\ref{fig:teacher_student_gap} tracks the teacher--student gap and its
rectification during training for Qwen2.5--3B and Qwen2.5--7B on ALFWorld,
WebShop, and Search-QA.

\subsection{Reward Score Curves}
\label{app:reward_curves}
\begin{figure}[p]
\centering
\includegraphics[width=\linewidth]{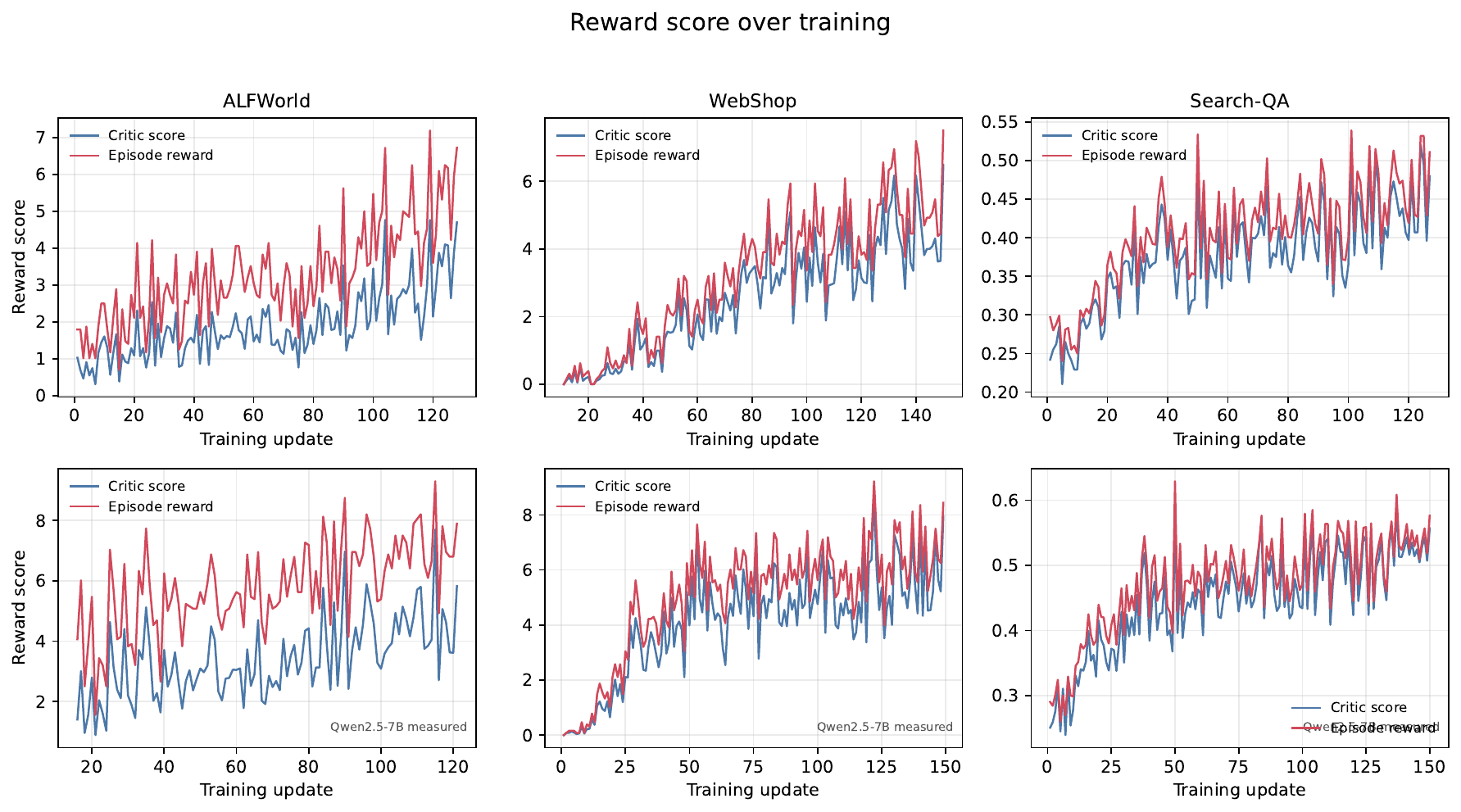}
\caption{\textbf{Critic scores and episode rewards during training.} Logged mean critic scores and episode rewards for Qwen2.5--3B/7B \textsc{AlignOPSD} runs on ALFWorld, WebShop, and Search-QA.}
\label{fig:appendix_reward_curves}
\end{figure}

Figure~\ref{fig:appendix_reward_curves} tracks the logged mean critic scores and
episode rewards during training for both backbones on all three benchmarks.

\clearpage

\subsection{Allocation Diagnostics}
\label{app:allocation_diagnostics}

\begin{figure}[htbp]
\centering
\includegraphics[width=\linewidth]{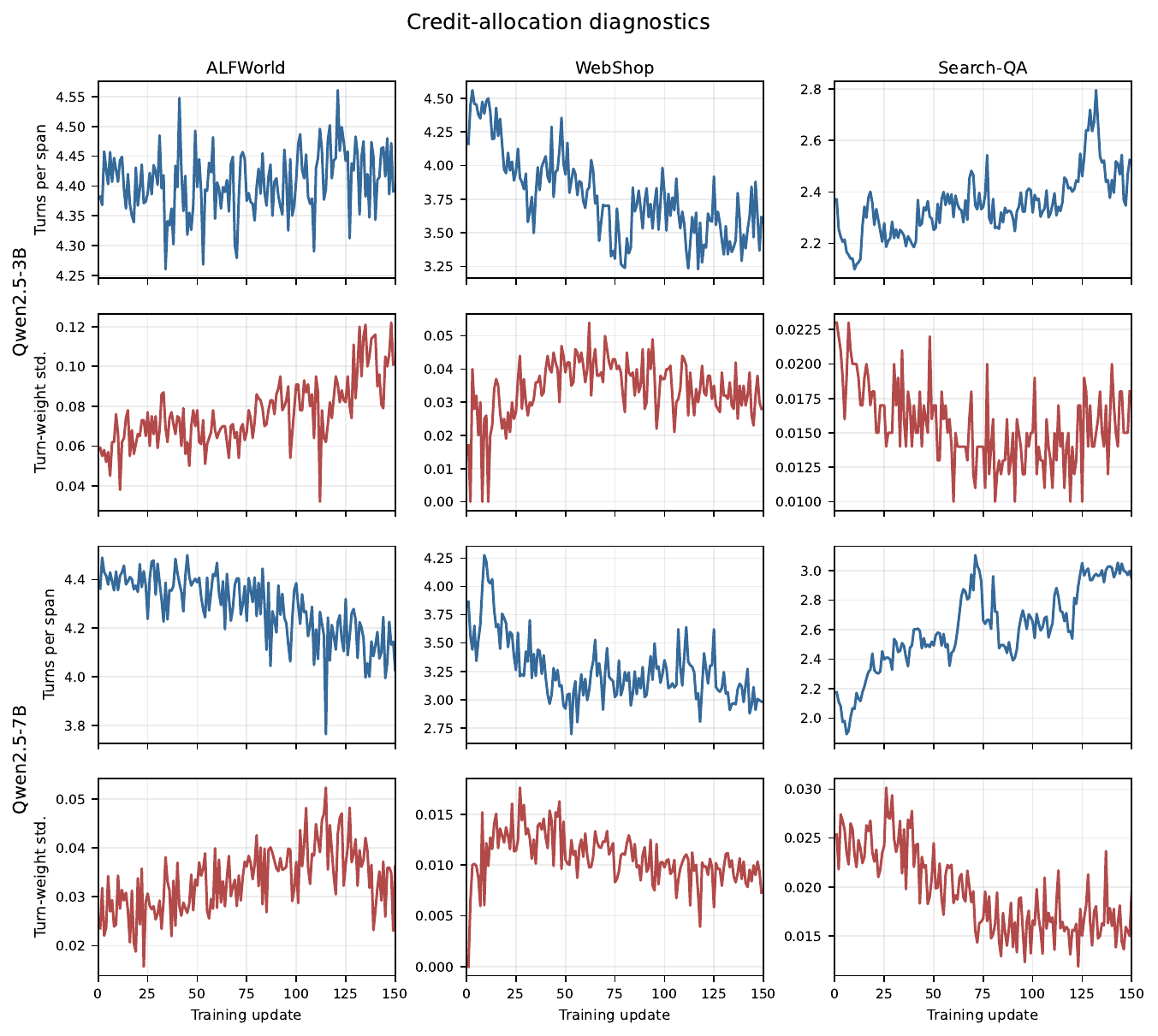}
\caption{\textbf{Allocation diagnostics during training.}
Columns show ALFWorld, WebShop, and Search-QA. For each backbone, the first row
reports turns per span and the second reports turn-weight standard deviation.
Y-axis labels appear only in the leftmost column and x-axis labels only in the
bottom row; each panel keeps its own y-axis ticks.}
\label{fig:allocation_diagnostics}
\end{figure}

Figure~\ref{fig:allocation_diagnostics} reports two allocation statistics:
turns per span measures the number of interaction turns grouped into each
allocated credit span, while turn-weight standard deviation measures the
concentration of credit across turns.

\clearpage
\subsection{Cross-Method Performance Traces}
\label{app:method_training_curves}

\begin{figure}[htbp]
\centering
\includegraphics[width=\linewidth]{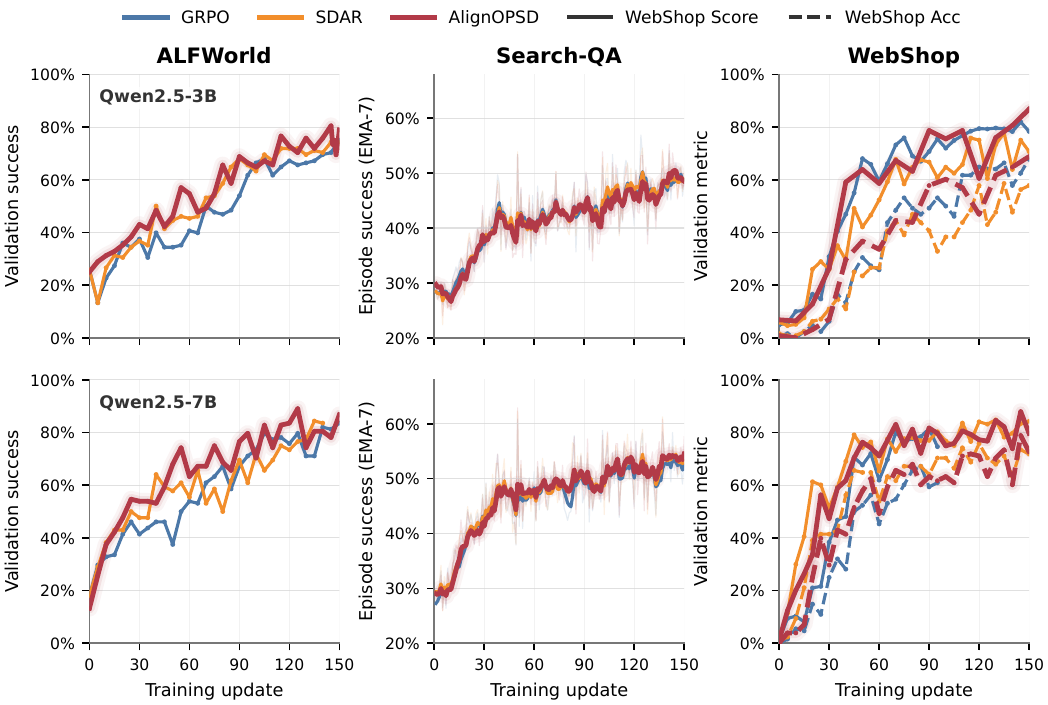}
\caption{\textbf{Cross-method performance traces.}
Rows correspond to Qwen2.5--3B and Qwen2.5--7B; columns correspond to ALFWorld,
Search-QA, and WebShop. ALFWorld uses validation success, Search-QA uses logged
episode success rate (EMA-7; faint lines show unsmoothed measurements), and
WebShop uses validation normalized score (solid) and exact success (dashed).
Each curve ends at its last logged update; no missing segment is extrapolated.}
\label{fig:method_training_curves}
\end{figure}

Figure~\ref{fig:method_training_curves} visualizes training-time performance
trajectories of GRPO, SDAR, and \textsc{AlignOPSD} across both backbones and all
three benchmarks. We report validation success on ALFWorld. Because Search-QA
does not provide comparable dense validation histories for all runs, we instead
show logged episode success rates smoothed with a seven-update exponential
moving average. WebShop reports both validation normalized score and exact
success, matching the two metrics in Table~\ref{tab:main_results}.

\end{document}